\documentclass{article}

\usepackage{amsmath}
\usepackage{graphics} 			
\usepackage{epsfig} 			
\usepackage{mathptmx} 			
\usepackage{times} 				
\usepackage{amsthm} 			
\usepackage{amsfonts}
\usepackage{amstext}
\usepackage{amssymb}
\usepackage{subfigure}
\usepackage[english]{babel}

\newtheorem*{theorem*}{Theorem}

\def\real{\mbox{\rm I\kern-0.18em R}}

\title{\LARGE \bf Synergy--based Hand Pose Sensing:\\ Reconstruction Enhancement\thanks{This work is supported by the European Commission under CP
grant no. 248587, THE Hand Embodied, within the FP7-ICT-2009-4-2-1
program Cognitive Systems and Robotics.}}

\author{Matteo Bianchi\thanks{The Interdept. Research Center ``En\-ri\-co Pi\-ag\-gio'', U\-ni\-ver\-si\-ty of Pisa, via Diotisalvi 2, 56100 Pisa, Italy. {\tt\footnotesize	 m.bianchi,p.salaris,bicchi@centropiaggio.unipi.it}}, \and Paolo Salaris\footnotemark[2], \and Antonio Bicchi\footnotemark[2] \thanks{Department of Advanced Robotics, Istituto Italiano di Tecnologia, via Morego, 30, 16163 Genova, Italy} }
\date{ }

\begin{document}

\maketitle
\thispagestyle{empty}
\pagestyle{empty}

\begin{abstract}
Low--cost sensing gloves for reconstruction posture provide measurements which are limited under several regards. They are generated through an imperfectly known model, are subject to noise, and may be less than the number of Degrees of Freedom (DoFs) of the hand. Under these conditions, direct reconstruction of the hand posture is an ill-posed problem, and performance can be very poor. This paper examines the problem of estimating the posture of a human hand using (low-cost) sensing gloves, and how to improve their performance by exploiting the knowledge on how humans most frequently use their hands. To increase the accuracy of pose reconstruction without modifying the glove hardware --- hence basically at no extra cost --- we propose to collect, organize, and exploit information on the probabilistic distribution of human hand poses in common tasks. We discuss how a database of such an {\em a priori} information can be built, represented in a hierarchy of correlation patterns or {\em postural synergies}, and fused with glove data in a consistent way, so as to provide a good hand pose reconstruction in spite of insufficient and inaccurate sensing data. Simulations and experiments on a low-cost glove are reported which demonstrate the effectiveness of the proposed techniques.
\end{abstract}

\section{Introduction}
\label{Int}

In recent years numerous studies have underlined the complex role of human hand in motor organization, with particular attention to grasping. It was shown that individuated finger motions were phylogenetically superimposed on basic grasping movements~\cite{Gordon}. Moreover, it is possible to individuate a reduced number of coordination patterns (\emph{synergies}) which constrain both joint motions and force exertions of multiple fingers~\cite{Schieber}. These constraints can be related to both biomechanics factors~\cite{Fahrer} and synchronization between different motor units~\cite{Kilbreath}. Coordination patterns were analyzed by means of multivariate statistical methods over a grasping data set, revealing that a limited amount of so-called \emph{eigenpostures} or principal components (PCs) \cite{Mason}, or otherwise ``statistically identified kinematic coordination patterns'' \cite{Schieber}, are sufficient to explain a great part of hand pose variability. In addition, a gradient in \emph{eigenpostures} was identified~\cite{Santelloart}, showing that lower order PCs take into account covariation patterns for metacarpophalangeal (MCP) and interphalangeal (IP) joints, which are mainly responsible for coarse hand opening and closing, while higher order PCs are used for fine hand shape adjustments.

These studies and results on human hand in grasping tasks suggest that there exist some inner hand representations of increasing complexity, which allow to reduce the number of Degrees of Freedom (DoFs) to be used according to the desired level of approximation. From a \emph{controllability} point of view, this idea was then adopted in robotics to define simplified approaches for the design and control of artificial hands \cite{Bicchi1,Brown}.

On the other hand, from the \emph{observability} point of view, this fact also suggests that it is possible to reduce the number of independent DoFs to be measured in order to obtained the hand pose estimation for a given level of approximation. An application of this concept was developed in \cite{Pratiart} for hand avatar animation. In this paper, we exploit the information embedded in a known grasp set, which expresses the postural constraints for multi-finger joints, to improve the reconstruction of the hand posture in static grasping tasks when only a limited and inaccurate number of measures are available by low-cost sensing gloves for gesture measurement~\cite{Sturman}, \cite{DiPietro}.
``Glove-based'' devices or ``sensing gloves'', i.e.~devices for hand pose reconstruction based on measurements of few geometric features of the hand, provide useful interfaces for human-machine and haptic interaction in many fields like, for example, virtual reality, musical performance, video games, teleoperation and robotics~\cite{Pao}. However, the widespread commercialization of electronic gloves imposes limits on the production costs in terms of the amount and the quality of the sensors adopted. As a consequence, the correctness of the hand pose reconstruction obtained by these devices might be compromised.

The objective of this paper and its companion~\cite{Bianchi_etalII} is to provide new tools to improve the design and performance of sensing gloves by exploiting the knowledge on how humans most frequently use their hands.

In this paper, partially based on~\cite{BianchiSal}, the aim is to provide hand pose estimation technique based on Bayes'inference which optimizes the performance for a given glove design. Two different approaches have been followed to achieve this goal. The first one solves a constrained optimization problem of multinormal probability density function (pdf), and it is mainly suited when accurately measured data is available. The second approach deals with noisy measured data and relies on classic Minimum Variance Estimation (MVE). To validate these methods we consider measurements from a set of grasp postures acquired with a low cost sensing glove and compare the achieved hand pose reconstruction with reference measures provided by a very accurate optical tracking system. Effects of noise are also taken into account and new simulations w.r.t.~\cite{BianchiSal} have been conducted. Statistical analyses of both experimental and simulation results demonstrate the effectiveness of the here proposed procedures.

In~\cite{Bianchi_etalII}, we consider the dual problem of optimal design of pose sensing gloves, based on a suitable cost function which combines the \emph{a priori} information and measures. Simulations are reported where the estimation procedures described in this paper are applied to the measures provided by the optimally designed glove. We first consider the case that individual sensing elements in the glove can be designed so as to measure a linear combination of joint angles, and provide, for given {\em a priori} information and fixed number of measurements, the optimal design minimizing in average the reconstruction error. We then discuss the case that only single-DoF sensors are used in the glove. Finally an hybrid design which combines continuous and discrete measures and expresses a trade-off between quality and feasibility is proposed.

\section{The hand posture estimation algorithm}
\label{hand}

Let us consider a set of measures $y\in\real^m$ given by a sensing glove. By using a $n$ degree of freedom kinematic hand model, we shall assume a linear relationship between joint variables $x\in\real^n$ and measurements $y$ given by
\begin{equation}
	\label{eq:MeasuresGlove}
		y = Hx+\nu\,,
\end{equation}
where $H\in\real^{m\times n}$ ($m<n$) is a full rank matrix which represents the relationship between measures and joint angles, and $v \in \real^m$ is a vector of measurement noise. The goal is to determine the hand posture, i.e.~the joint angles $x$, by using a set of measures $y$ whose number is lower than the number of DoFs describing the kinematic hand model in use. Equation~\eqref{eq:MeasuresGlove} represents a system where there are fewer equations than unknowns and hence is compatible with an infinite number of solutions, described e.g.~as
\begin{equation}
	x = H^\dagger y+N_h\xi\,,
	\label{eq:Estimation}
\end{equation}
where $H^\dagger$ is the pseudo-inverse of matrix $H$, $N_h$ is the null space basis of matrix $H$ and $\xi\in\real^{(n-m)}$ is a free vector of parameters.
Among these possible solutions, the least-squared solution resulting from the pseudo-inverse of matrix $H$ for system~\eqref{eq:MeasuresGlove} is a vector of minimum Euclidean norm given by
\begin{equation}
	\label{eq:Pseudoinverse}
	\hat x=H^\dagger y\,.
\end{equation}

However, the hand pose reconstruction resulting from~\eqref{eq:Pseudoinverse} can be very far from the real one. The purpose of this paper is to improve on the accuracy of the pose reconstruction, choosing, among the possible solution to~\eqref{eq:Estimation}, the most likely hand pose. The basic idea is to exploit the fact that human hands, although very complex and possibly different in size and shape, share many commonalities in how they are shaped and used in frequent everyday tasks. Indeed, studies on the human hand in grasping tasks showed that finger motions are strongly correlated according to some coordination patterns referred to as synergies in~\cite{Santelloart}. 

In this paper, we show how to improve hand pose reconstruction by exploiting the {\em a priori} information obtained by collecting a large number $N$ of grasp postures $x_i$, consisting of $n$ DoFs, into a matrix  $X\in\real^{n\times N}$. This information can be summarized in a covariance matrix $P_o\in\real^{n\times n}$, which is a symmetric matrix computed as
\[
P_o=\frac{(X-\bar{x})(X-\bar{x})^{T}}{N-1}\,,
\]
where $\bar{x}$ is a matrix $n\times N$ whose columns contain the mean values for each joint angle arranged in vector $\mu_o\in\real^{n}$. 

\subsection{Probability Density Function Maximization}

In this section, we initially consider the case that measurement noise is negligible. The hand pose estimation can be improved w.r.t.~that obtained by~\eqref{eq:Pseudoinverse}, by exploiting the \emph{a priori} information, that we will assume to be a multivariate normal distribution, on a set of grasping poses built beforehand and embedded in the covariance matrix $P_o$. 
The best estimation of the hand posture is given by choosing as optimality criterion the maximization of the \emph{probability density function} (pdf) of a multivariate normal distribution, expressed by (\cite{Tarantola})
\begin{equation}
f(x)=\frac{1}{\sqrt{2\pi\|P_o\|}}\exp\left\{-\frac{1}{2}(x -\mu_o)^{T}P_o^{-1}(x -\mu_o) \right\}\,.
\label{eq:p}
\end{equation}
This is equivalent to solving the following optimal problem:
\begin{equation}
\begin{cases}
\hat{x}=\underset{\hat{x}}{\arg \min}\; \frac{1}{2}(x -\mu_o)^{T}P_o^{-1}(x -\mu_o)\\
\text{Subject to}\quad y = Hx\,.
\label{eq:M22}
\end{cases}
\end{equation}

It is interesting to give a geometrical interpretation of the cost function in~\eqref{eq:M22}, which expresses the square of the Mahalanobis distance~\cite{Mahalanobis}. The concept of Mahalanobis distance, which takes into account data covariance structure, is widely exploited in statistics, e.g.~in PC Analysis, mainly for outlier detection \cite{Hawkins1980}. Accordingly, to assess if a test point belongs to a known data set, whose distribution defines an hyper-ellipsoid, we take into account both its closeness to the centroid of data set and the direction of the test point w.r.t.~the centroid itself. In other words, the more samples are distributed along this direction, the more probably the test point belongs to the data set even if it is further from the center.

Taking into account~\eqref{eq:Estimation}, the optimal problem defined in~\eqref{eq:M22} becomes
\begin{equation}
\begin{cases}
\hat{\xi}=\underset{\hat{\xi}}{\arg \min}(H^\dagger y + N_h\xi - \mu_o)^{T}P_o^{-1}(H^\dagger y + N_h\xi - \mu_o)\\
\text{Subject to}\quad y = Hx\,.
\label{eq:M22bis}
\end{cases}
\end{equation}
By using classic optimization procedures we obtain
$
\hat{\xi}=(N_h^{T}P_o^{-1}N_h)^{-1}N_h^{T}P_o^{-1}(\mu_o - H^\dagger y)
$
and, substituting in~\eqref{eq:Estimation}, after some algebras, the estimation of the hand joint angles is
\begin{align}
\hat{x} = [I - N_h(N_h^{T}P_o^{-1}N_h)^{-1} & N_h^{T}P_o^{-1}]H^\dagger y +\nonumber\\
& +N_h(N_h^{T}P_o^{-1}N_h)^{-1}N_h^{T}P_o^{-1}\mu_o\,.
\label{eq:L}
\end{align}

Problem~\eqref{eq:M22} can be also solved through the method of Lagrange multipliers. Introduce a new variable $\lambda\in\real^m$ and consider
\begin{equation}
L = \frac{1}{2}(x -\mu_o)^{T}P_o^{-1}(x -\mu_o) + \lambda^{T}(Hx-y)\,.
\label{eq:LagrangeFunction}
\end{equation}
By imposing $\frac{\partial L}{\partial x} = \frac{\partial L}{\partial \lambda} = 0$, we have
\begin{equation}
\hat{x} = \mu_o - P_oH^{T}(HP_oH^{T})^{-1}(H\mu_o-y)\,.
\label{eq:L2}
\end{equation}
This solution can be easily shown to be equivalent to~\eqref{eq:L}.

Finally, it is interesting to observe that the least-squared and pdf maximization methods have a direct application in case of only single-DoF sensors are used in the devices (discrete sensing gloves). In this case, $H$ is a selection matrix whose rows are vectors of the canonical basis in $\real^n$ and the least-squared solution is simply given as $\hat x = H^T y$. In order to improve the hand pose reconstruction by the {\em a priori} information, it is possible to easily maximize $E[x|y]$ in terms of multinormal conditional distribution~\cite{Hardle}. Indeed, in this case vector $y$ defines a precise subset of the state variables, being $X_1$, whose values are known by means the measurement process, while $X_2$ indicates the rest of state variables to be estimated. This definition allows to partition the \emph{a priori} covariance matrix as
\begin{equation}
\left(\frac{X_1}{X_2}\right) \Longrightarrow P_o = \left(\frac{P_{o11}|P_{o12}}{P_{o21}|P_{o22}}\right)
\label{eq:M25}
\end{equation}
as well as the \emph{a priori} mean $\mu_o = (\mu_{o1}|\mu_{o2})$. The estimation of $X_2$ is easily derived as
\begin{equation}
\hat{X}_2 = E[X_2|X_1 = y] = \mu_{o2} + P_{o21}P_{o11}^{-1}(y-\mu_{o1})\,.
\label{eq:L3}
\end{equation}

\subsection{Minimum Variance Estimation}

Results in previous section are valid in the condition of $\nu\approx0$. When noise is not negligible, the role of {\em a priori} is more emphasized.

In this section we propose an algorithm based on the Minimum Variance Estimation (MVE) technique. This method minimizes a cost functional which expresses the weighted Euclidean norm of deviations, i.e.~cost functional $J = \int_X (\hat{x}-x)^{T}S(\hat{x}-x)dx$, where $S$ is an arbitrary, semidefinite positive matrix.

Under the hypothesis that $\nu$ has zero mean and Gaussian distribution with covariance matrix $R$, we get the solution for the minimization of $J$ as $\hat{x} = E[x|y]$, where $E[x|y]$ represents the \emph{a posteriori} pdf expectation value.
The estimation $\hat{x}$ can be obtained as in~\cite{Gelb} by
\begin{equation}
\hat{x} = (P_o^{-1}+H^{T}R^{-1}H)^{-1}(H^{T}R^{-1}y + P_o^{-1}\mu_o)\,,
\label{eq:MAP}
\end{equation}
where matrix $P_p = (P_o^{-1}+H^{T}R^{-1}H)^{-1}$ is the \emph{a posteriori} covariance matrix, which has to be minimized to increase information about the system. This result represents a very common procedure in applied optimal estimation when there is redundant sensor information. In under-determined problems, it is only thanks to the {\em a priori} information, represented by $P_o$ and $\mu_o$, that equation~\eqref{eq:MAP} can be applied (indeed, $H^TR^{-1}H$ is not invertible).

When $R$ tends to assume very small values, the solution described in equation \eqref{eq:MAP} might encounter numerical problems.
However, by using the Sherman-Morrison-Woodbury formulae,
\begin{align}
	\label{eq:Equality1}
	(P_o^{-1}+H^TR^{-1}H)^{-1} &= P_o-P_oH^T(HP_oH^T+R)^{-1}HP_o\\
	(P_o^{-1}+H^TR^{-1}H)^{-1} & H^TR^{-1} = P_oH^T(HP_oH^T+R)^{-1}\,,
	\label{eq:Equality2}
\end{align}
equation~\eqref{eq:MAP} can be rewritten as
\begin{equation}
\label{eq:MVE2}
\hat{x} = \mu_o - P_oH^{T}(HP_oH^{T}+R)^{-1}(H\mu_o-y)\,,
\end{equation}
and the {\em a posteriori} covariance matrix becomes $P_p=P_o-P_oH^T(HP_oH^T+R)^{-1}HP_o$ (cf.~\eqref{eq:Equality1}).
By placing $R=0$ in~\eqref{eq:MVE2}, we obtain equation~\eqref{eq:L2} and the \emph{a posteriori} covariance matrix becomes
\begin{align}
P_p &= P_o-P_oH^T(HP_oH^T)^{-1}HP_o\,.
\label{eq:AposterioriMP}
\end{align}

Notice that probability density function maximization approach is a peculiar case of the here described MVE technique. For this reason in the following sections we will always refer to the reconstruction technique as MVE for both noise--free and noisy measures and we will use~\eqref{eq:MVE2} with $R=0$ or $R\neq 0$, respectively.

\section{Model and Data capture}
\label{Section:mod}
\setcounter{figure}{0}
\begin{figure}[!t]
	\centering
	\begin{tabular}[c]{c}
		\includegraphics[width=0.4\columnwidth]{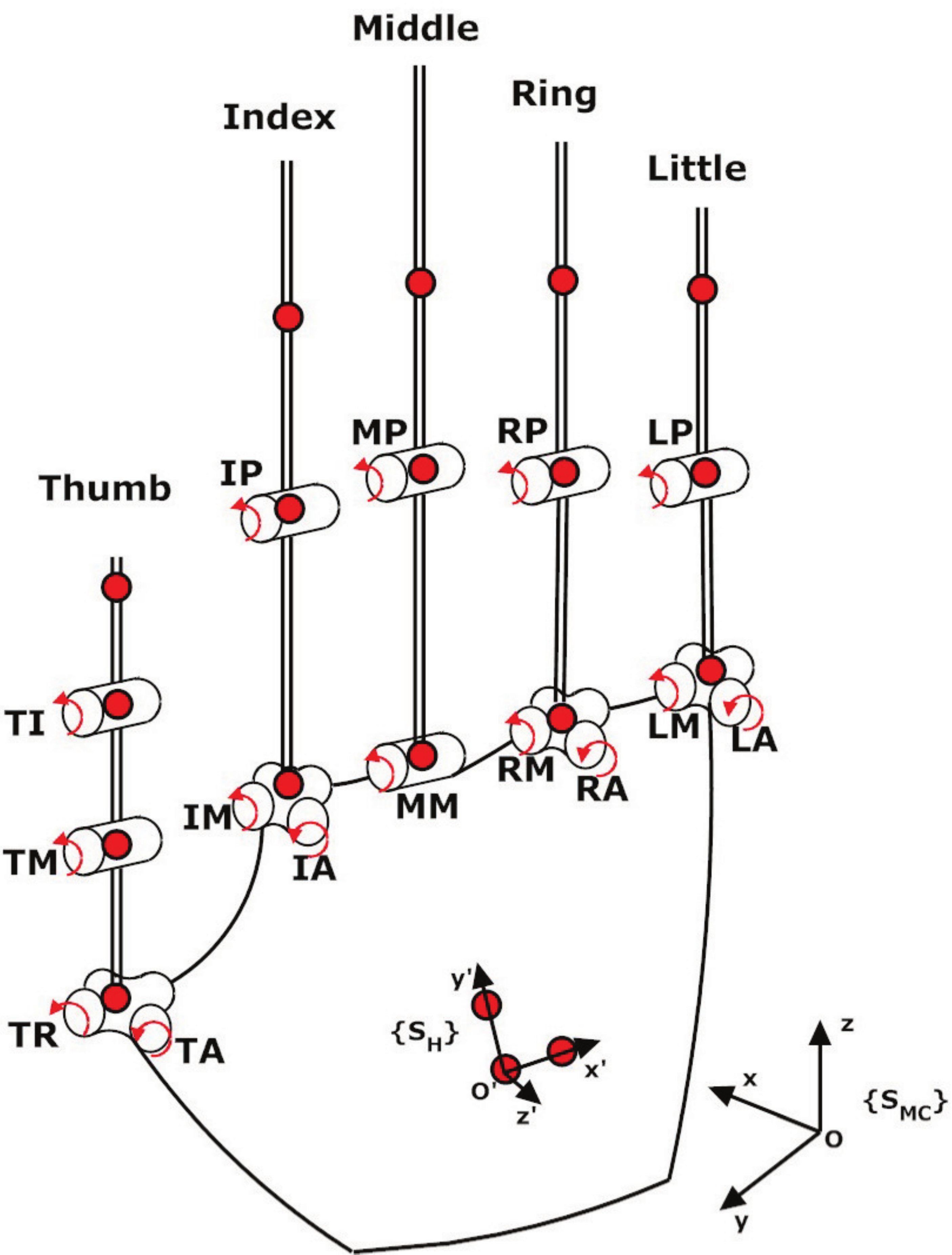}
	\end{tabular}
	\hspace{-.4cm}
	\renewcommand{\arraystretch}{1.05}
	\small{
	\begin{tabular}[c]{|c|c|}
		\hline
		\textbf{DoFs} & \textbf{Description}\\
		\hline
		TA & Thumb Abduction\\
		\hline
		TR & Thumb Rotation\\
		\hline
		TM & Thumb Metacarpal\\
		\hline
		TI & Thumb Interphalangeal\\
		\hline
		IA & Index Abduction\\
		\hline
		IM & Index Metacarpal\\
		\hline
		IP & Index Proximal\\
		\hline
		MM & Middle Metacarpal\\
		\hline
		MP & Middle Proximal\\
		\hline
		RA & Ring Abduction\\
		\hline
		RM & Ring Metacarpal\\
		\hline
		RP & Ring Proximal\\
		\hline
		LA & Little abduction\\
		\hline
		LM & Little Metacarpal\\
		\hline
		LP & Little Proximal\\
		\hline
	\end{tabular}}
	\caption{Kinematic model of the hand with 15 DoFs. Markers are reported as red spheres.}
\label{fig:KinMod}
\end{figure}

Without loss of generality, for hand pose reconstruction we adopt the 15 DoFs model also used in \cite{Santelloart, Gabicciniart} and reported in figure~\ref{fig:KinMod}. A large number of static grasp positions were collected using 19 active markers and an optical motion capture system (Phase Space, San Leandro, CA - USA). More specifically, all the grasps of the 57 imagined objects described in \cite{Santelloart} were performed twice by subject AT (M,26), in order to define a set of 114 \emph{a priori} data. Moreover, 54 grasp poses of a wide range of different imagined objects were executed by subject LC (M,26)~\footnote{These hand posture acquisitions are available at http://handcorpus.org/}. These data is recorded in parallel with the sensing glove and the Phase Space system, to achieve both a glove calibration and reliable reference values for whole hand configurations. Indeed, we can consider the processed hand poses acquired with Phase Space as a good approximation of real hand positions, given the high accuracy provided by this optical system to detect markers (the amount of static marker jitter is inferior than 0.5 mm, usually 0.1 mm) and assuming a linear  correlation (due to skin stretch) between marker motion around the axes of rotation of the joint and the movement of the joint itself \cite{Zhang}. Since the sensing glove perfectly adapts to subject hand shape when it is worn, the latter assumption is still reasonable also in this case.
None of the subjects had physical limitations that would affect the experimental outcomes. Data collection from subjects in this study was approved by the University of Pisa Institutional Review Board.
\begin{figure}[!h]
\centering
\includegraphics[width=0.41\columnwidth]{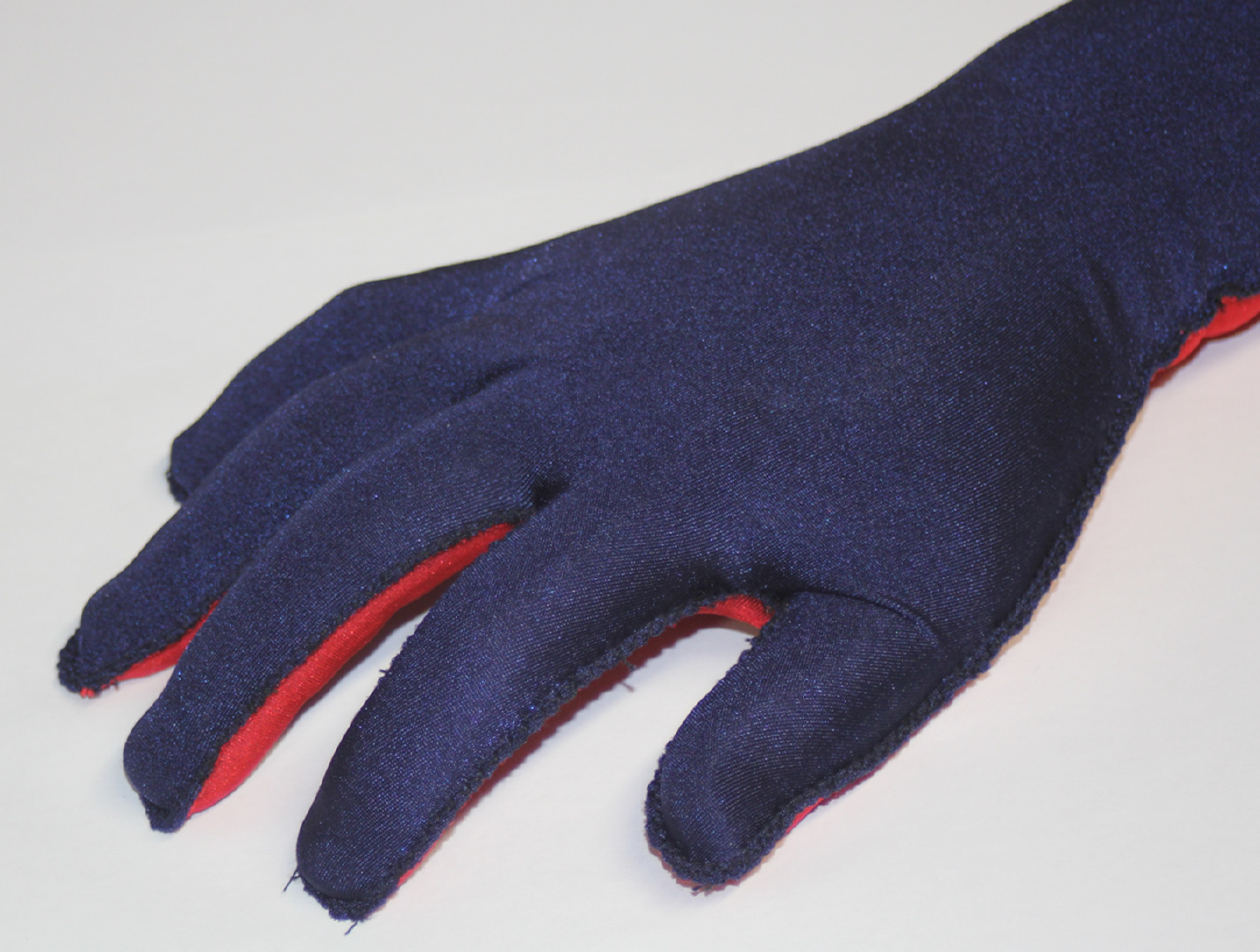}\quad
\includegraphics[width=0.47\columnwidth]{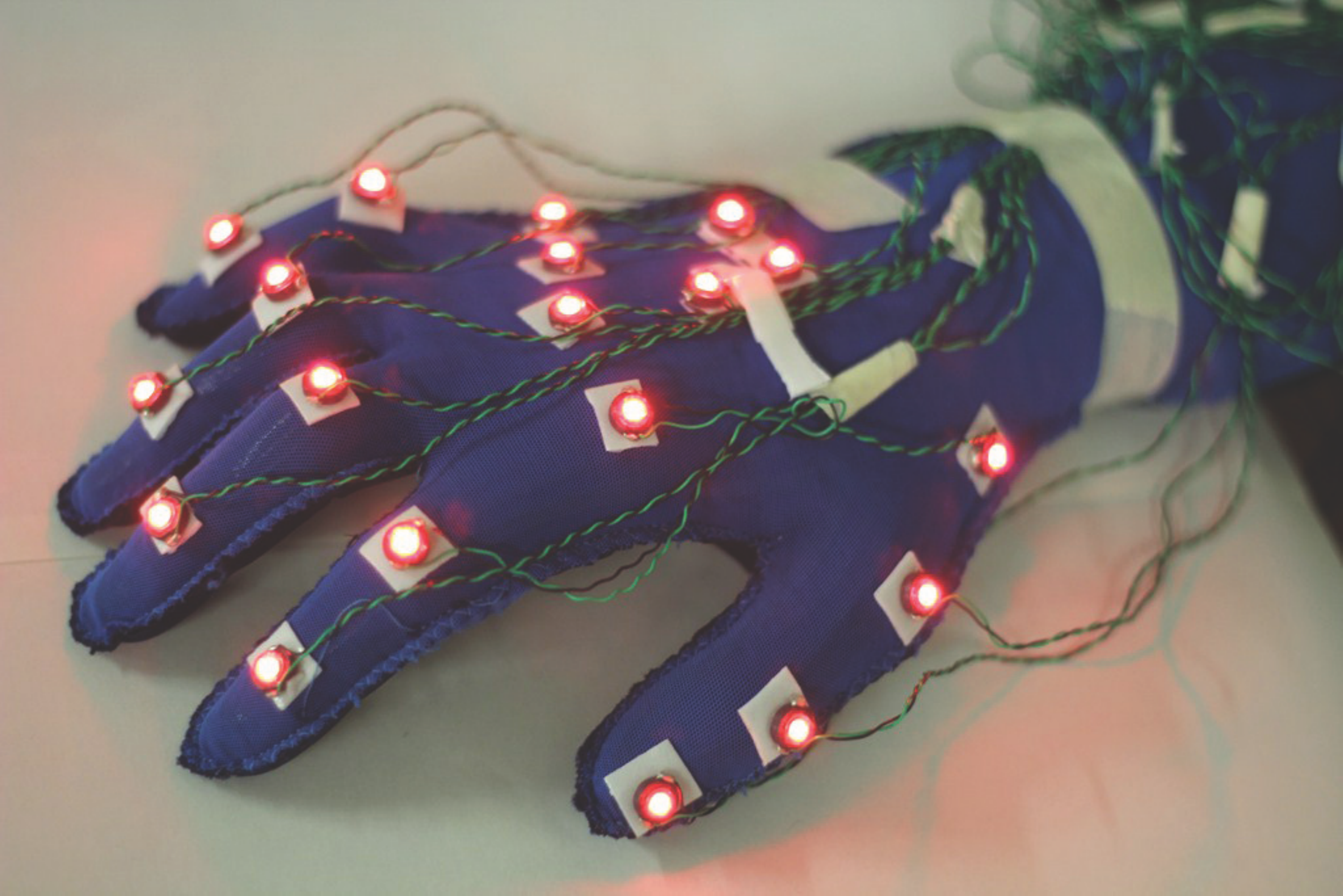}
\caption{The sensing glove (on the left) and sensing glove with added markers (on the right).}
\label{fig:guantol}
\end{figure}
Markers are placed on the glove referring to \cite{Qiushiart}, see figure~\ref{fig:guantol}. Four markers are used for the thumb and three markers for each of the rest of the fingers. Additional three markers placed on the dorsal surface of the palm defines a local reference system $S_H$. Markers are sampled at 480 Hz and their positions are given referring to the global reference system $S_{MC}$, as it is defined during the calibration phase of the acquisition system (cf.~figure~\ref{fig:KinMod}).

Joint angles are computed w.r.t.~$S_H$ by means the \emph{ikine} function of Matlab Kinematic Toolbox. This function implements an iterative algorithm of kinematic inversion, which has been suitably modified by adapting computational tolerance to guarantee numerical convergence. A moving average filter is exploited for data pre-filtering, thus enhancing Signal Noise Ratio (SNR). As a preliminary phase, the hand is posed in a reference position, where fingers flexion-extension is nearly zero, and phalanx length and eventual offset angles are computed.

Normality assumption on the acquired \emph{a priori} set is tested by means of a Q-Q plot-based graphical method for multidimensional variables~\cite{Chambers83,Holg}. The quantile plot is usually obtained by plotting the ordered estimated Mahalanobis measures against the chi-square distribution quantiles. If normality is met, the graph should display a fairly straight line on the diagonal (i.e.~$45^{\circ}$ slope line). In our case, the linear fitting with straight $45^{\circ}$ slope line provides an adjusted r-squared coefficient of 0.6.
This result suggests that the normality assumption is reasonable even if not fully met. However, the Gauss-Markov theorem~\cite{Raobook73} ensures, that the MVE is the Best Linear Unbiased Estimate (BLUE) in the minimum-variance sense even for non-Gaussian a priori distributions~\cite{Bicchi_Can1}.
In addition, central limit theorem~\cite{Hardle} can guarantee, to some extents, the application of MVE method to cases that depart from the strict linear-Gaussian hypothesis for \emph{a priori} distribution (and noise distribution as well).

\section{Simulation Results}
\label{Results}

We simulate an ideal glove which is able to measure only metacarpal joints by using the acquisitions obtained with Phase Space. The measurement matrix for this simulated glove will be referred as $H_s$. Estimation results obtained with our algorithms are compared with the corresponding reference values. An additional random Gaussian noise with standard deviation of $7{^\circ}$ is considered on each measurement. This value is chosen in a cautionary manner, based on data about common technologies and tools used to measure hand joints positions~\cite{Simone}. More specifically, this value expresses the reliability threshold of manual goniometry with skilled therapists in measures for rehabilitation procedures~\cite{Wiseetal90}.

The estimation performance is evaluated in terms of estimation errors.
Pose estimation errors (i.e. the mean of DoF absolute estimation errors computed for each pose), and DoF absolute estimation errors are considered and averaged over all the number of reconstructed poses. We perform these two types of analysis in order to furnish a more clear result comprehension. Indeed, pose estimation errors provide an useful but only global indication about the technique outcomes, potentially leading to some biased observations. For example, we might obtain a hand pose reconstruction with all the fingers in a ``slightly right'' position producing the same average error of a hand pose reconstruction with all the fingers but one in the right position and the one mispositioned very distant from the real one. Therefore, to overcome this limitation we also analyze each DoF estimation accuracy. In addition, some reconstructed poses are displayed w.r.t the reference ones, to provide a qualitative representation mainly focused on reconstruction \emph{likelihood} exhibited by reconstructed poses with common grasp postures.
Statistical differences between estimated pose and joint errors obtained with above described techniques are computed by using classic tools, after having tested for normality and homogeneity of variances assumption on samples (through Lilliefors' composite goodness-of-fit test and Levene's test, respectively). Standard two-tailed t-test (hereinafter referred as $T_{eq}$~) is used in case of both the assumptions are met, a modified two-tailed T-test is exploited (Behrens-Fisher problem, using Satterthwaite's approximation for the effective degrees of freedom, hereinafter referred as $T_{neq}$~) when variance assumption is not verified and finally a non parametric test is adopted for the comparison (Mann-Whitney U-test, hereinafter referred as $U$~) when normality hypothesis fails. Significance level of 5\% is assumed and p-values less than $10^{-4}$ are posed equal to zero.

In case of noise free measurements, mean absolute pose error obtained with MVE is $6.69\pm2.38^{\circ}$,
while with Pinv it is equal to  $13.89\pm3.09^{\circ}$, with observed  statistical difference between the two methods ($\text{p}\simeq0$, $T_{eq}$~).
What is noticeable is that MVE provides a better pose estimate than the one obtained using Pinv in terms of both mean pose absolute estimate error and considering maximum absolute pose estimation error (MVE: $13.18^{\circ}$ vs.~Pinv: $20.82^{\circ}$).

In case of noisy measurements, mean absolute pose estimation error with MVE is $8.52\pm2.86^{\circ}$, while with Pinv we get $15.71\pm3.08^{\circ}$. Also in this case statistical difference is observed between MVE and Pinv ($\text{p}\simeq 0$, $T_{neq}$~).
Notice that MVE still provides the best pose estimate and the smallest pose absolute maximum error (MVE: ($17.14^{\circ}$ vs.~Pinv: $23.39^{\circ}$).

\begin{table}[h!]
\centering
\includegraphics[width=0.7\columnwidth]{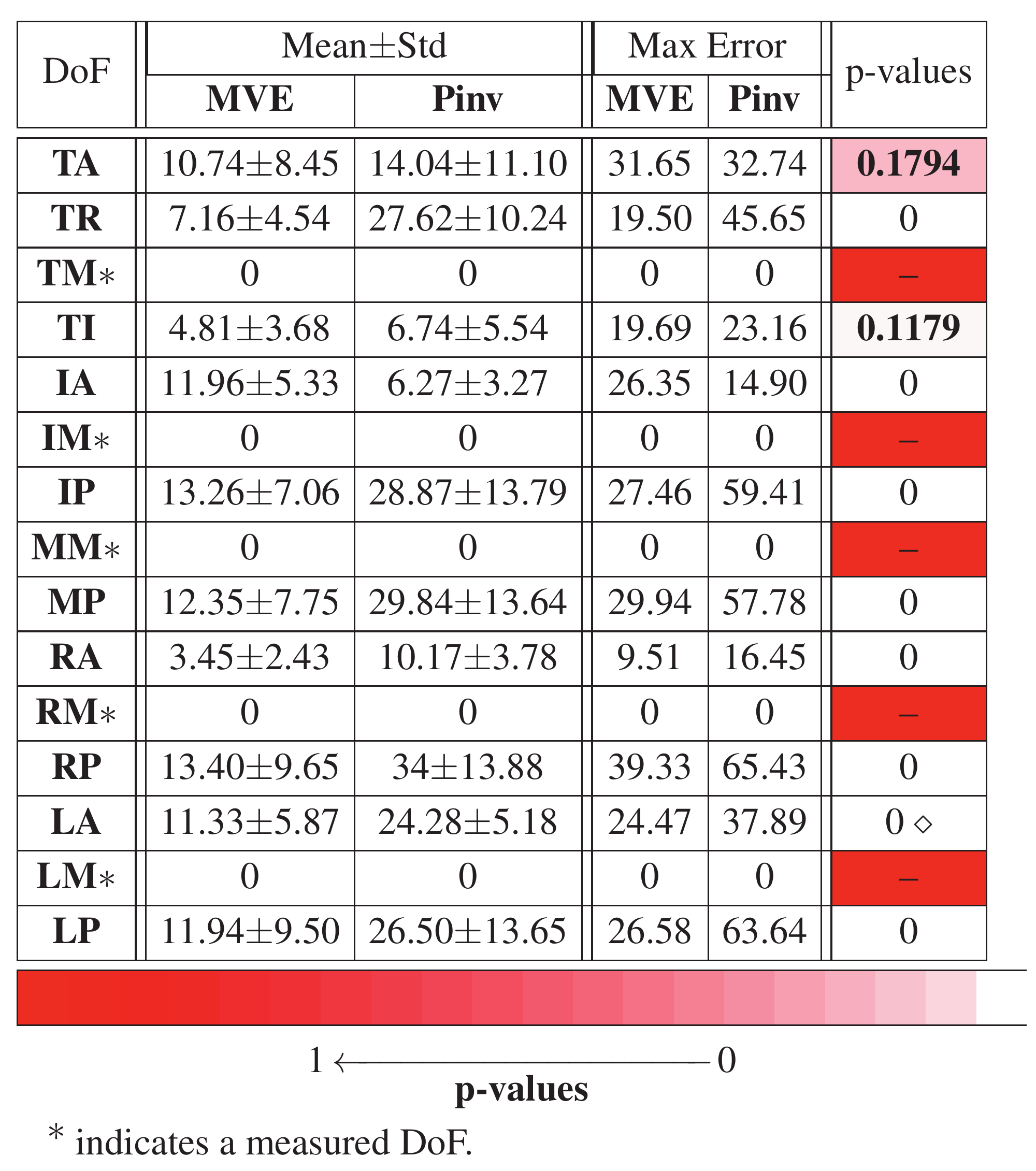}
\caption{Average estimation errors and standard deviations for each DoF $[^{\circ}]$ for the simulated acquisitions without noise. MVE and Pinv methods are considered. Maximum errors are also reported as well as p-values from the evaluation of DoF estimation errors between MVE and Pinv. A color map describing p-values is also added to simplify result visualization. $\diamond$ indicates  that $T_{eq}$ test is exploited for the comparison. $\ddagger$ indicates a $T_{neq}$ test. When no symbol appears near the tabulated values, it means that $U$ test is used. $\bold{Bold}$ value indicates no statistical difference between the two methods under analysis at 5\% significance level. When the difference is significative, values are reported with a $10^{-4}$ precision. p-values less than $10^{-4}$ are considered equal to zero.}
\label{tab:ZeroArt}
\end{table}
\begin{table}[t!]
\centering
\includegraphics[width=0.7\columnwidth]{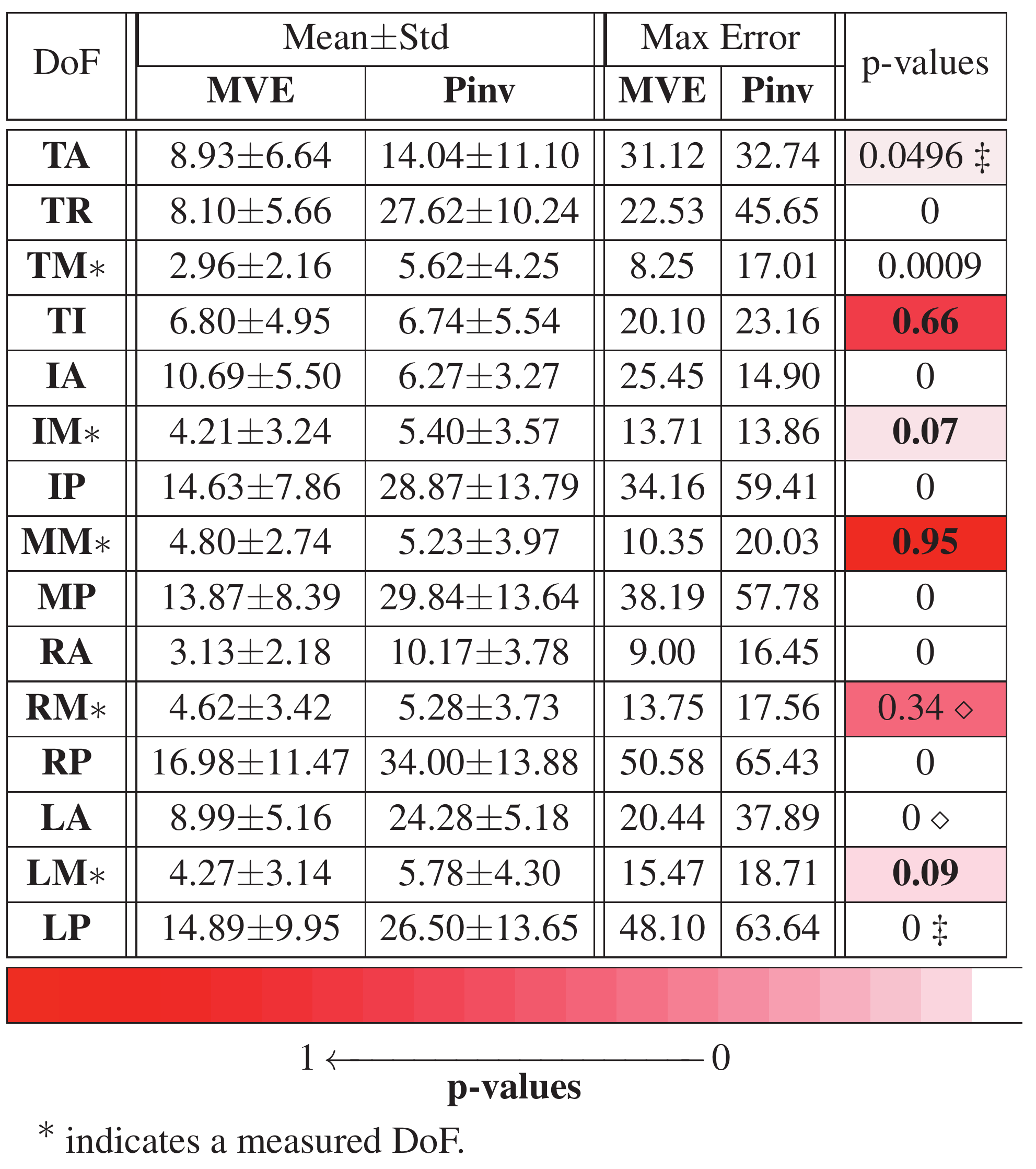}
\caption{Average estimation errors and standard deviations for each DoF $[^{\circ}]$ for the simulated acquisitions with noise. MVE and Pinv methods are considered. Maximum errors are also reported as well as p-values from the evaluation of DoF estimation errors between MVE and Pinv. A color map describing p-values is also added to simplify result visualization. $\diamond$ indicates  that $T_{eq}$ test is exploited for the comparison. $\ddagger$ indicates a $T_{neq}$ test. When no symbol appears near the tabulated values, it means that $U$ test is used. $\bold{Bold}$ value indicates no statistical difference between the two methods under analysis at 5\% significance level. When the difference is significative, values are reported with a $10^{-4}$ precision. p-values less than $10^{-4}$ are considered equal to zero.}
\label{tab:Giuntisimart}
\end{table}

In table~\ref{tab:ZeroArt} absolute average estimation errors for each DoF with their corresponding standard deviations are reported for MVE and Pinv procedures. Noise-free measures are considered. Significant statistical differences between the two techniques are found considering estimation errors for all DoFs, except for those directly measured and for TA and TI. The latter ones refer to thumb finger; this fact might be partially explained by the difficulties in modeling thumb phalanges under a kinematic point of view. MVE exhibits an estimation performance in terms of mean error which is better or not statistically different from the one achieved by Pinv, except for IA DoF; however, in this case the difference between the mean errors for the two methods is the smallest (less than $6^{\circ}$) among all the differences computed for the significantly different estimated DoFs. MVE provides the smallest maximum errors except for IA DoF; however the difference with maximum error obtained using Pinv is less than $12^\circ$. This difficulty in estimating IA DoF might be partially explained in terms of the variability in controlling index abduction, which can lead to slightly different position w.r.t the ones resulting from the \emph{a priori} set. In table~\ref{tab:Giuntisimart} values of each DoF estimation absolute error averaged over all poses, with their corresponding standard deviations, are reported in case of noise. Maximum errors are calculated and statistical significance in result comparison for each DoF estimation, between the aforementioned techniques, is indicated in table~\ref{tab:Giuntisimart}. Notice that MVE furnishes the best performance with average estimation errors which are always inferior or not statistically different from the ones obtained using Pinv algorithm, except for IA DoF for which Pinv produces the smallest average estimation error. However, the difference between IA mean errors calculated with the two procedures is less than $3^{\circ}$. No statistically significant difference are found between MVE and Pinv for TI, IM, MM, RM and LM DoFs.
Notice that the DoFs for which no statistical difference is observed between MVE and Pinv are DoFs directly measured or they refer to thumb finger phalanx. This fact may be partially explained in terms of the difficulties in thumb phalanges modeling under a kinematic point of view as previously mentioned.
Considering maximum errors, MVE still exhibits the best results, except for IA DoF.

\begin{figure*}[t!]
\begin{center}
	\renewcommand{\arraystretch}{1.5}
\begin{tabular}[c]{c}
Real Hand Postures\\
\end{tabular}
\begin{tabular}[c]{p{0.3cm}|p{2.4cm}|p{2.4cm}|p{2.4cm}|p{2.4cm}|}
\cline{2-5}
 &
\includegraphics[width=0.21\textwidth]{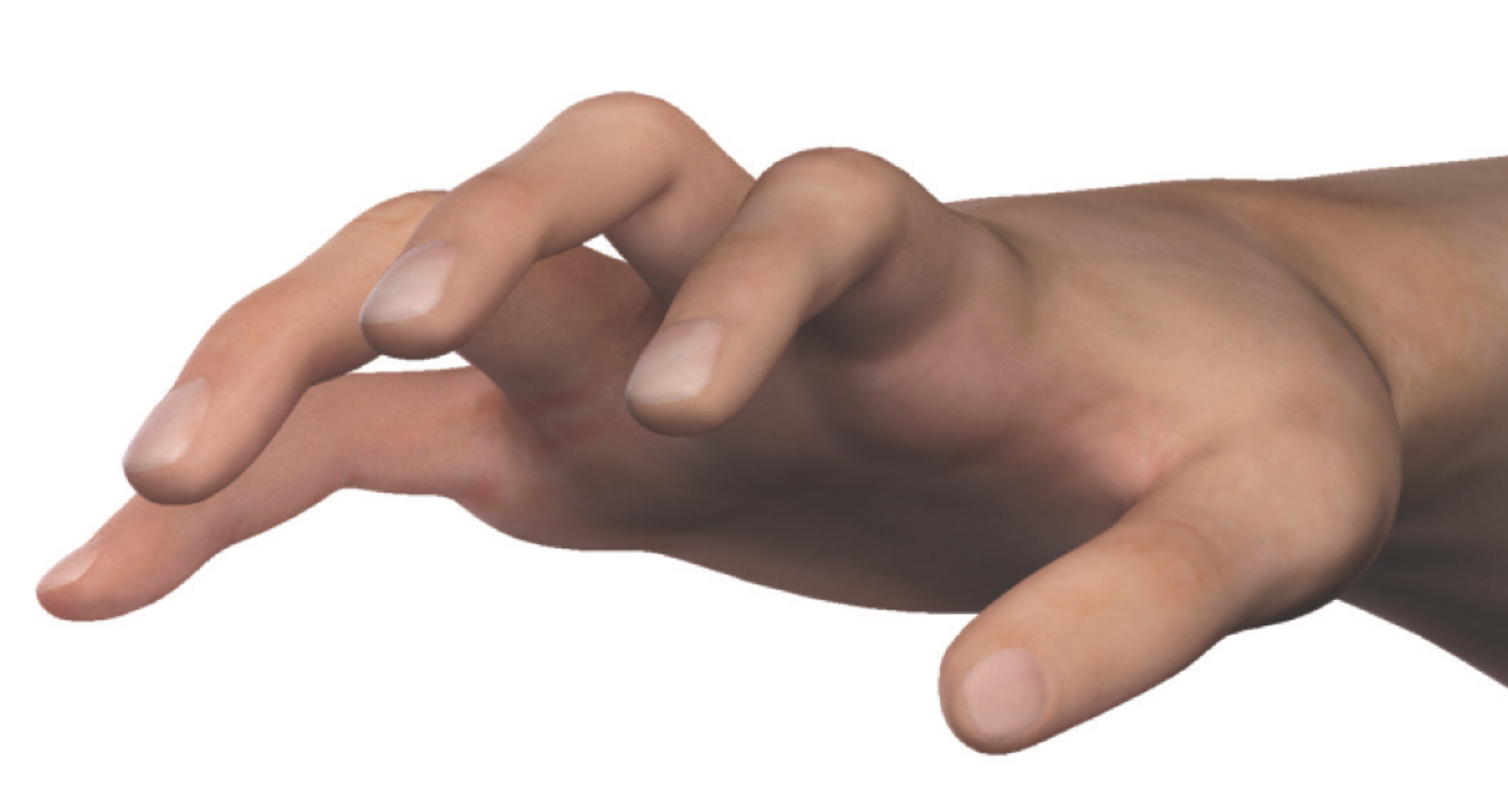} &
\includegraphics[width=0.185\textwidth]{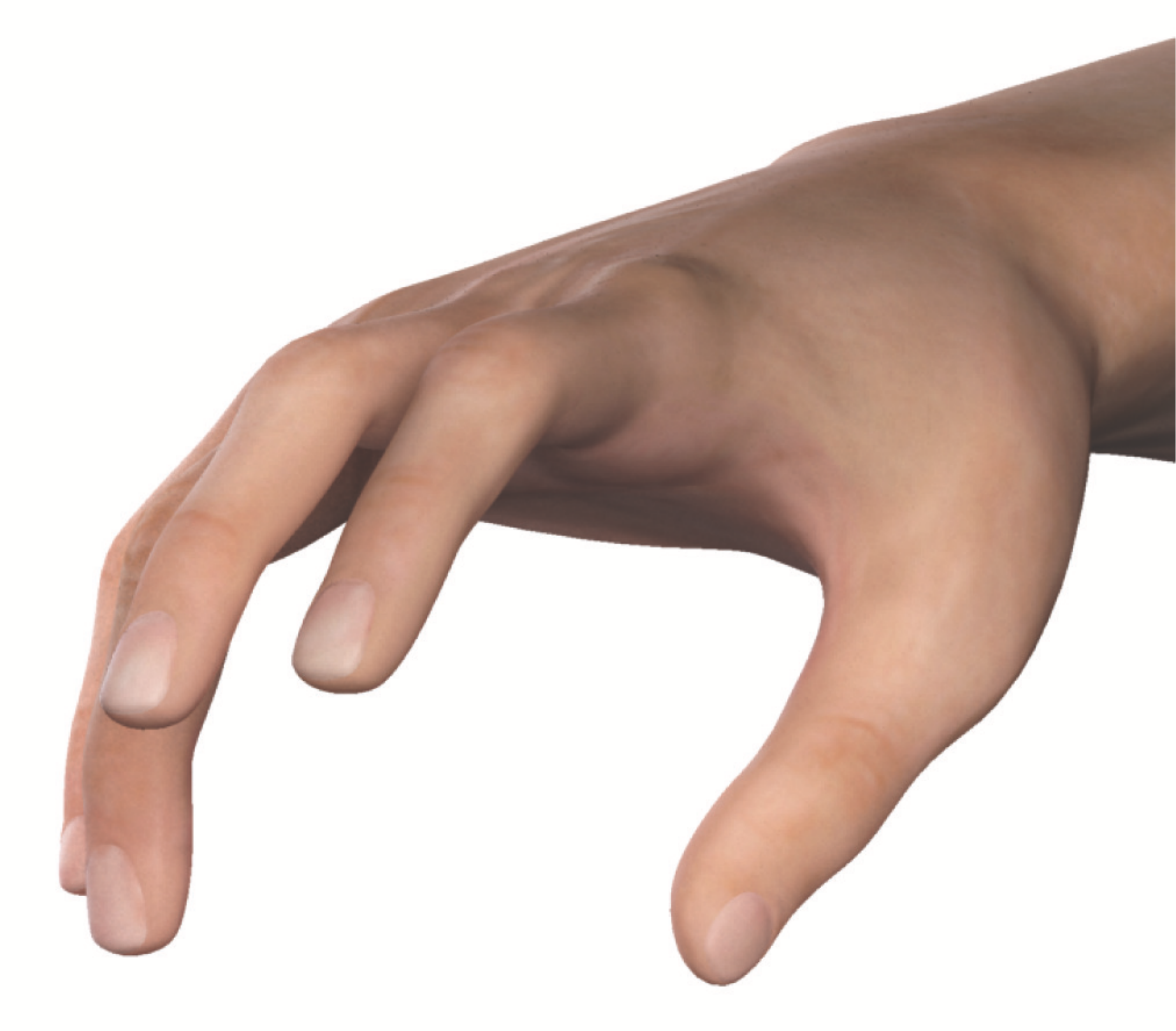} &
\includegraphics[width=0.165\textwidth]{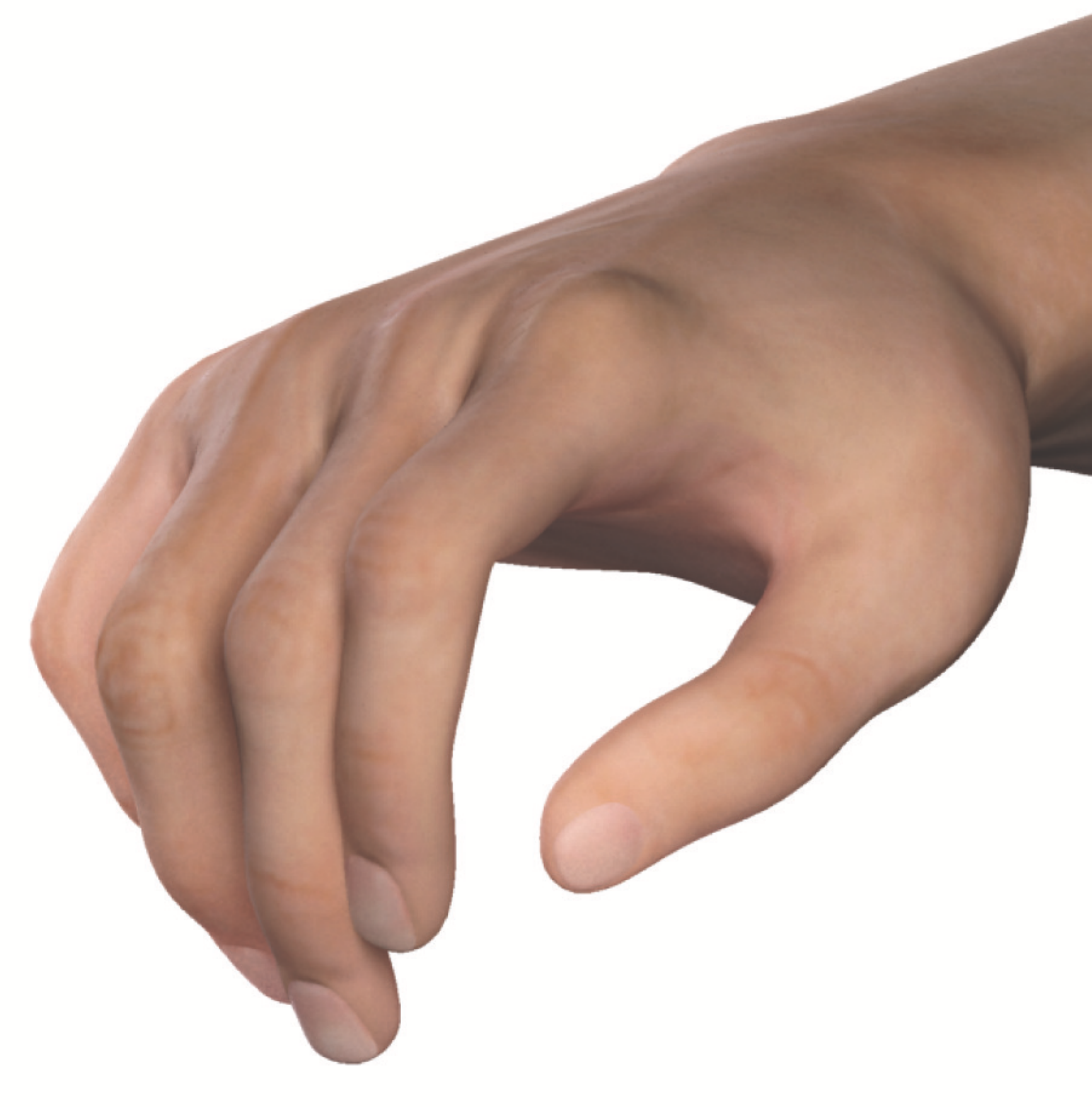} &
\includegraphics[width=0.2\textwidth]{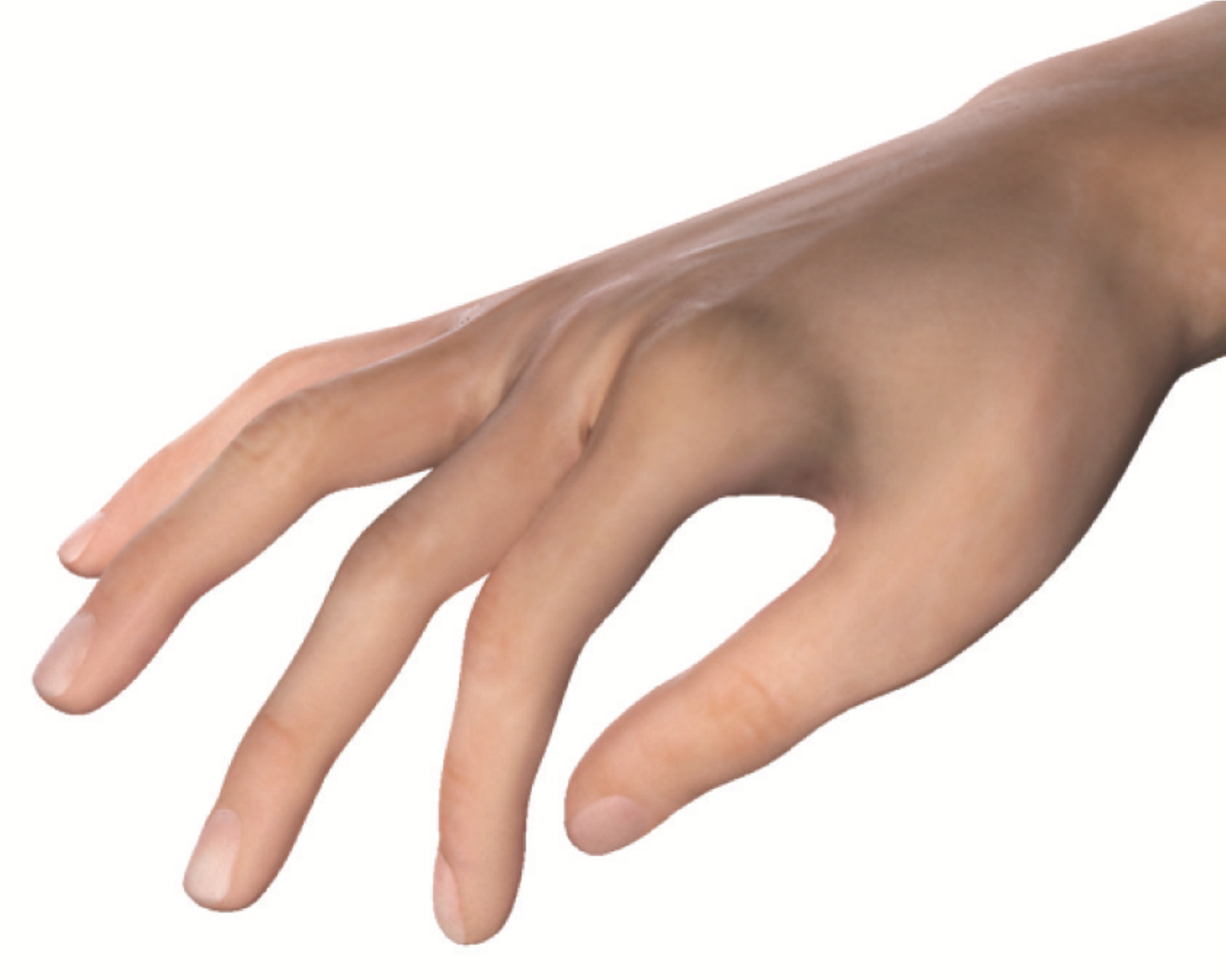}\\
\cline{2-5}
\cline{2-5}
\end{tabular}
\begin{tabular}[c]{c}
Posture estimation by noiseless measures\\
\end{tabular}
\begin{tabular}[c]{|p{0.3cm}|p{2.4cm}|p{2.4cm}|p{2.4cm}|p{2.4cm}|}
\hline
{\rotatebox{90}{\mbox{Pinv}}} &
\includegraphics[width=0.22\textwidth]{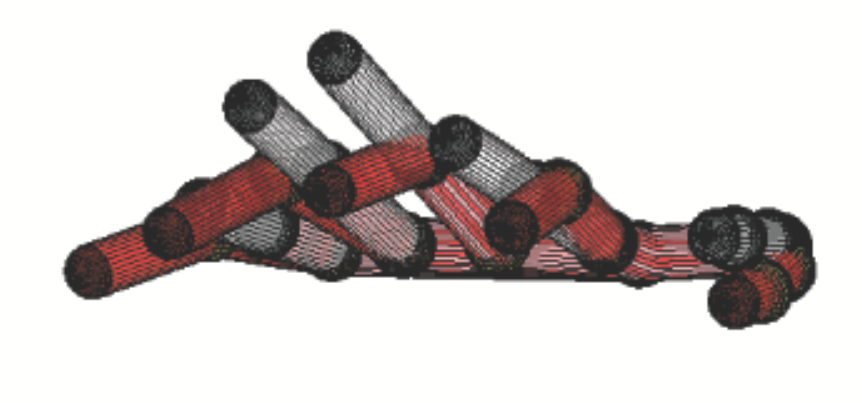} &
\includegraphics[width=0.2\textwidth]{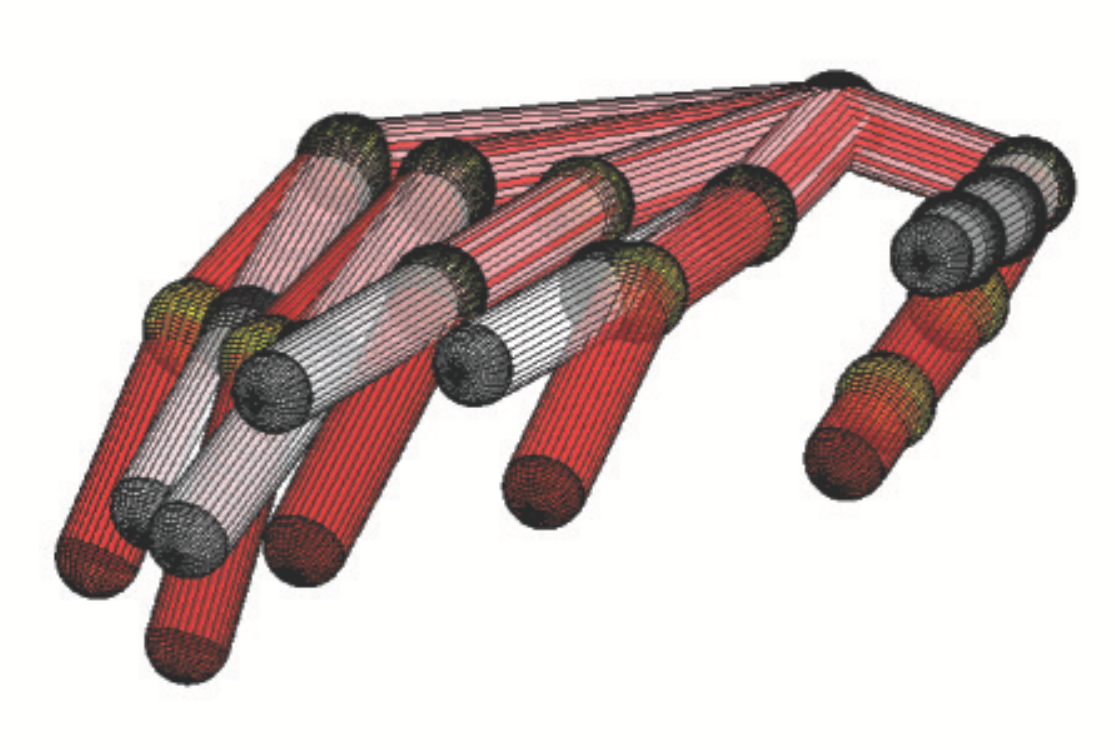} &
\includegraphics[width=0.18\textwidth]{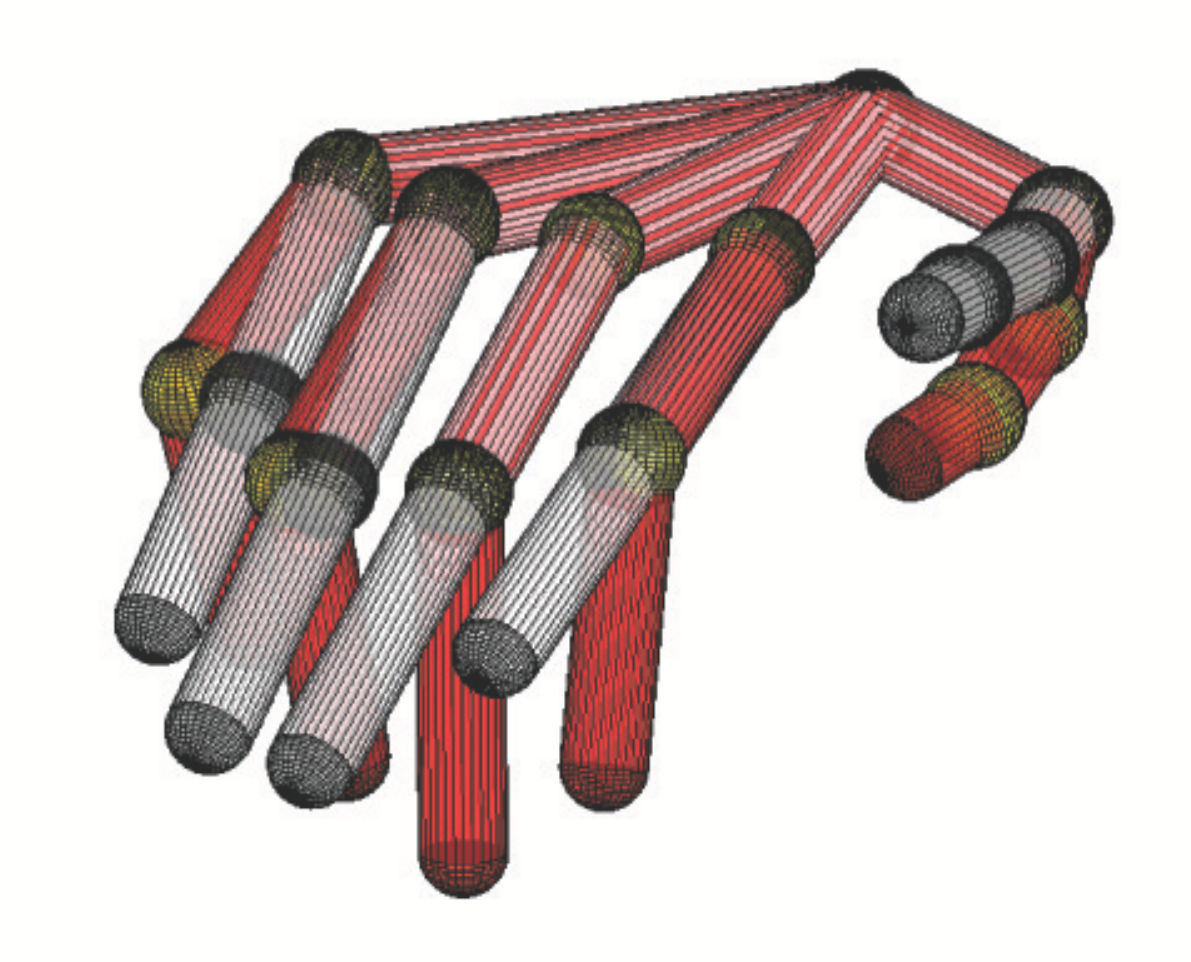} &
\includegraphics[width=0.2\textwidth]{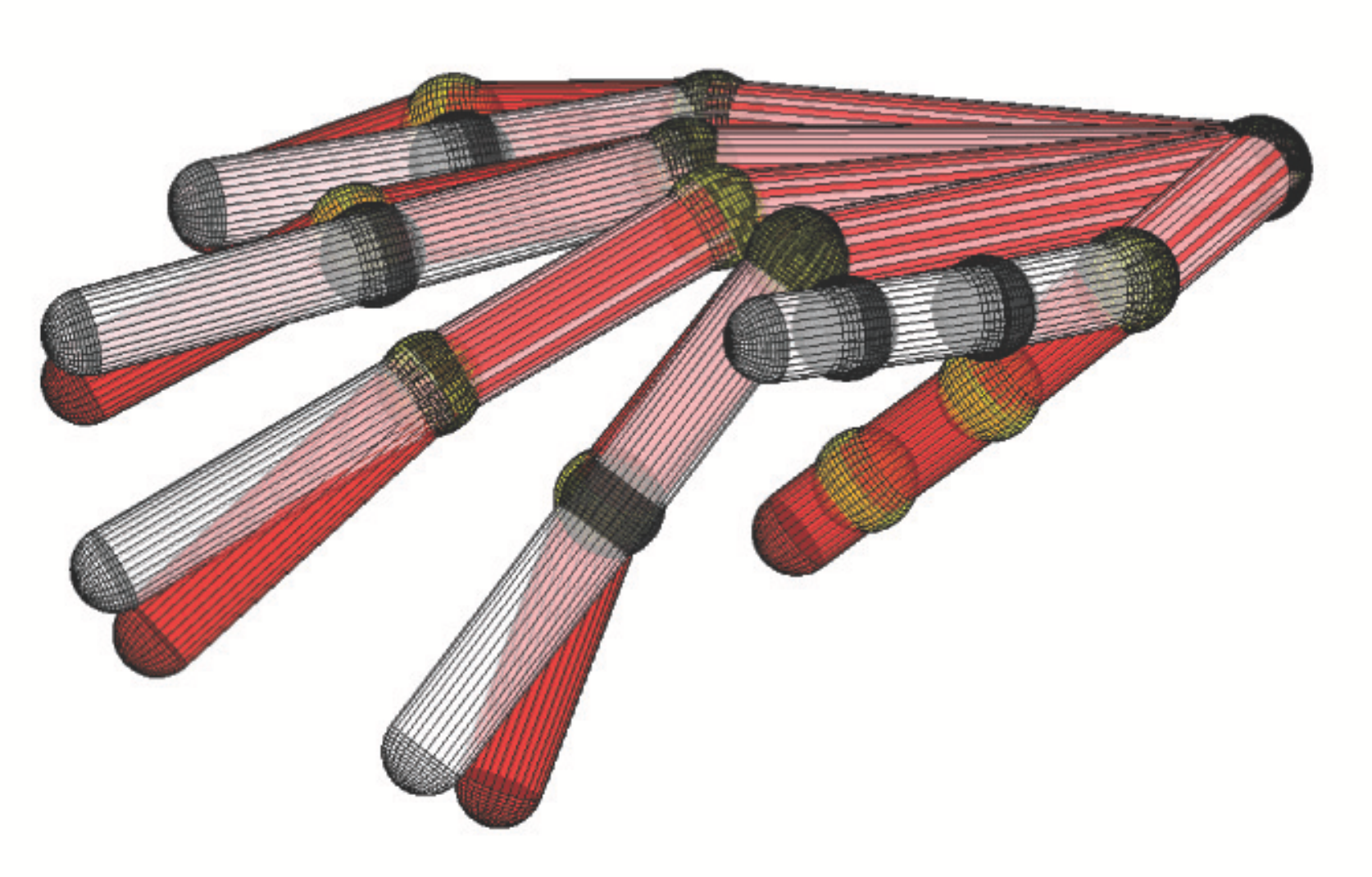}\\
\hline
{\rotatebox{90}{\mbox{MVE}}} &
\includegraphics[width=0.21\textwidth]{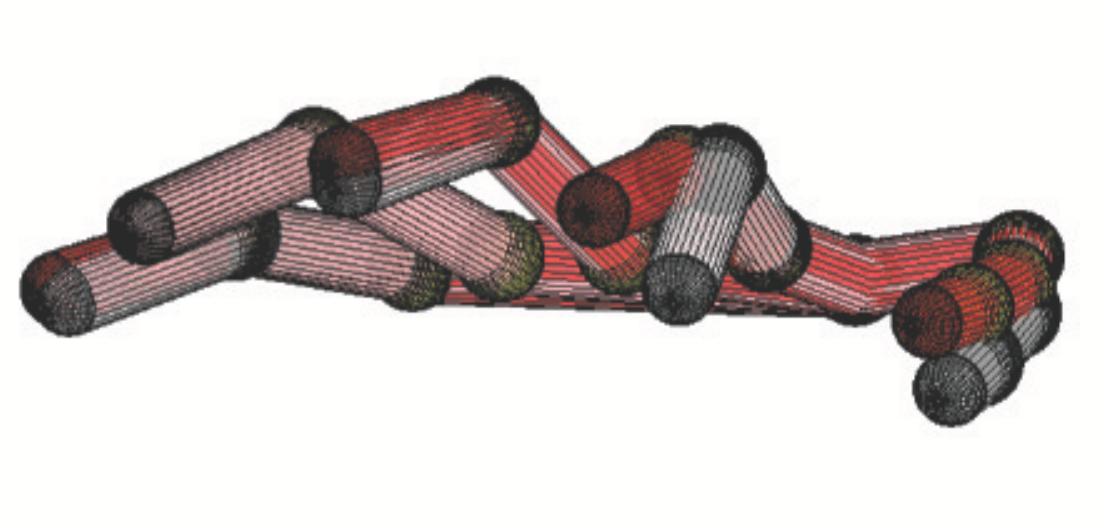} &
\includegraphics[width=0.2\textwidth]{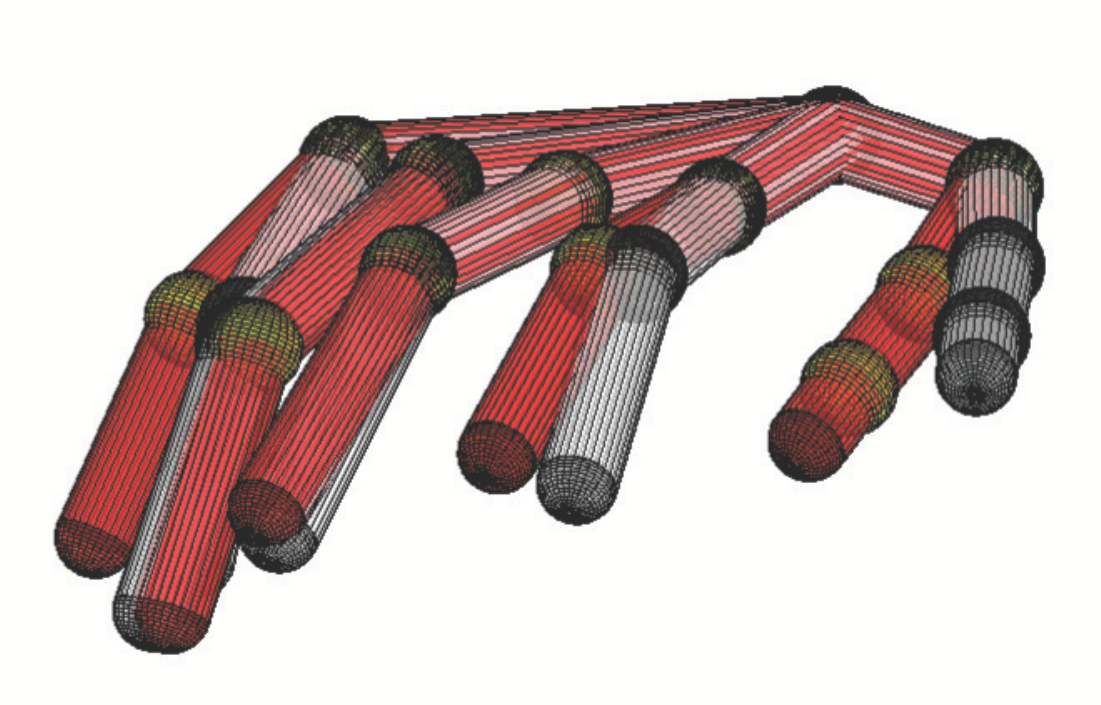} &
\includegraphics[width=0.175\textwidth]{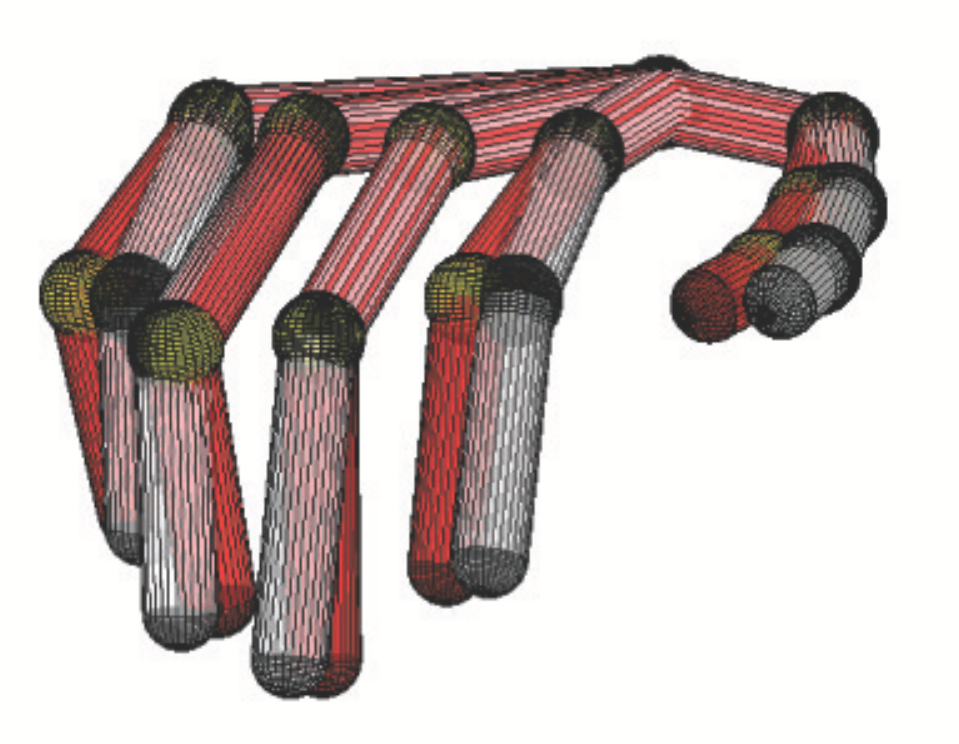} &
\includegraphics[width=0.19\textwidth]{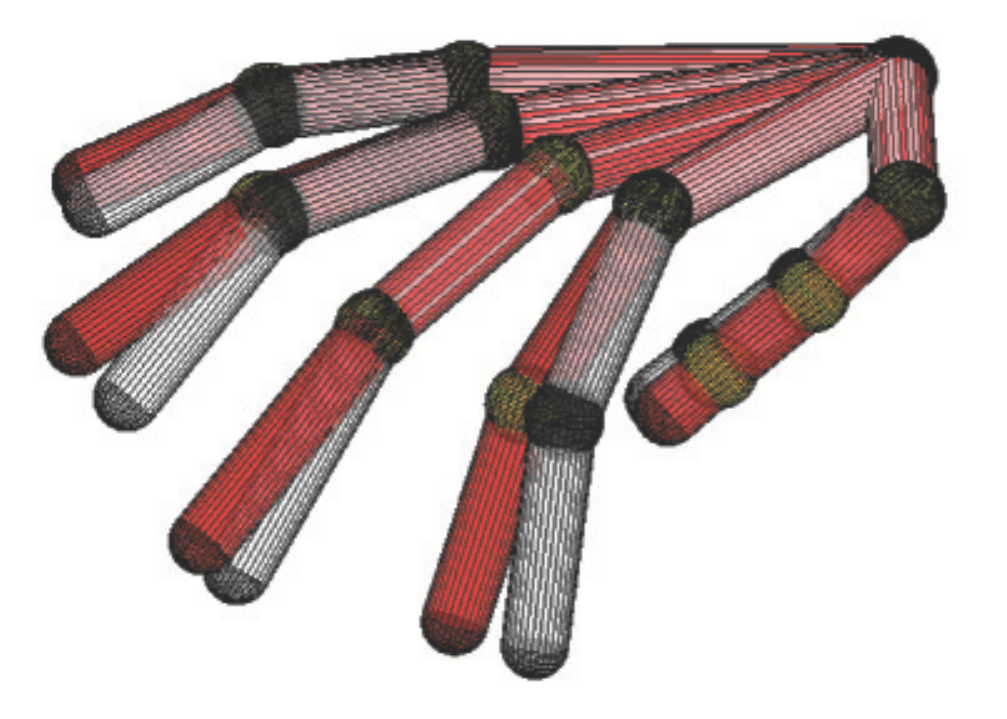}\\
\hline
\hline
\end{tabular}
\begin{tabular}[c]{c}
Posture estimation using measures affected by Gaussian noise\\
\end{tabular}
\begin{tabular}[c]{|p{0.3cm}|p{2.4cm}|p{2.4cm}|p{2.4cm}|p{2.4cm}|}
\hline
{\rotatebox{90}{\mbox{Pinv}}} &
\includegraphics[width=0.21\textwidth]{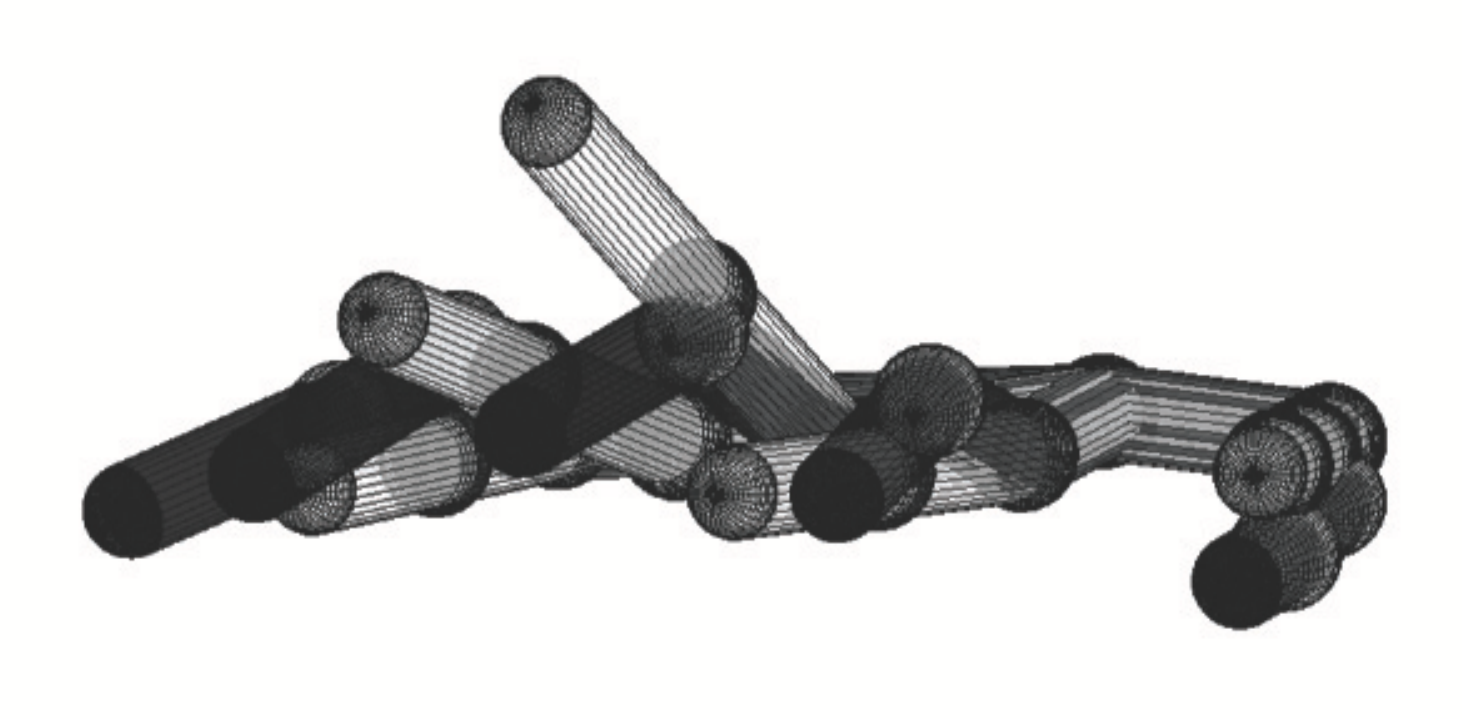} &
\includegraphics[width=0.21\textwidth]{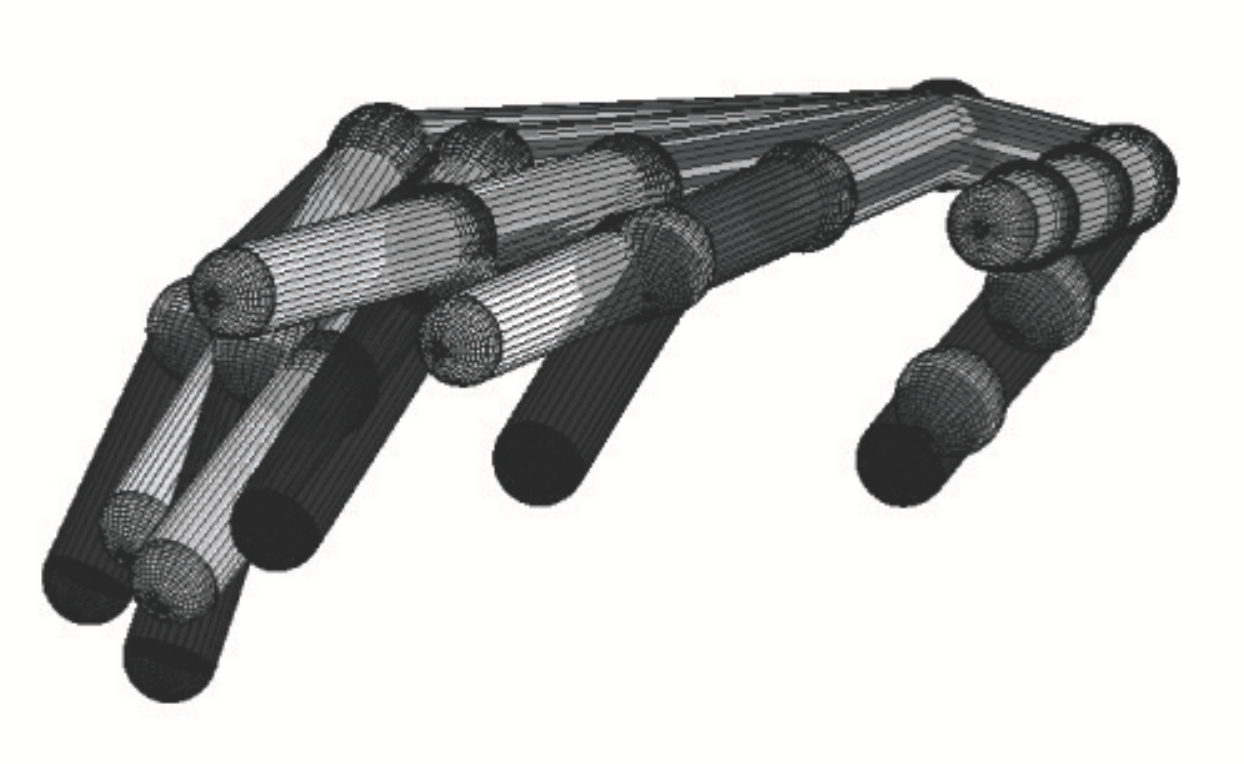} &
\includegraphics[width=0.18\textwidth]{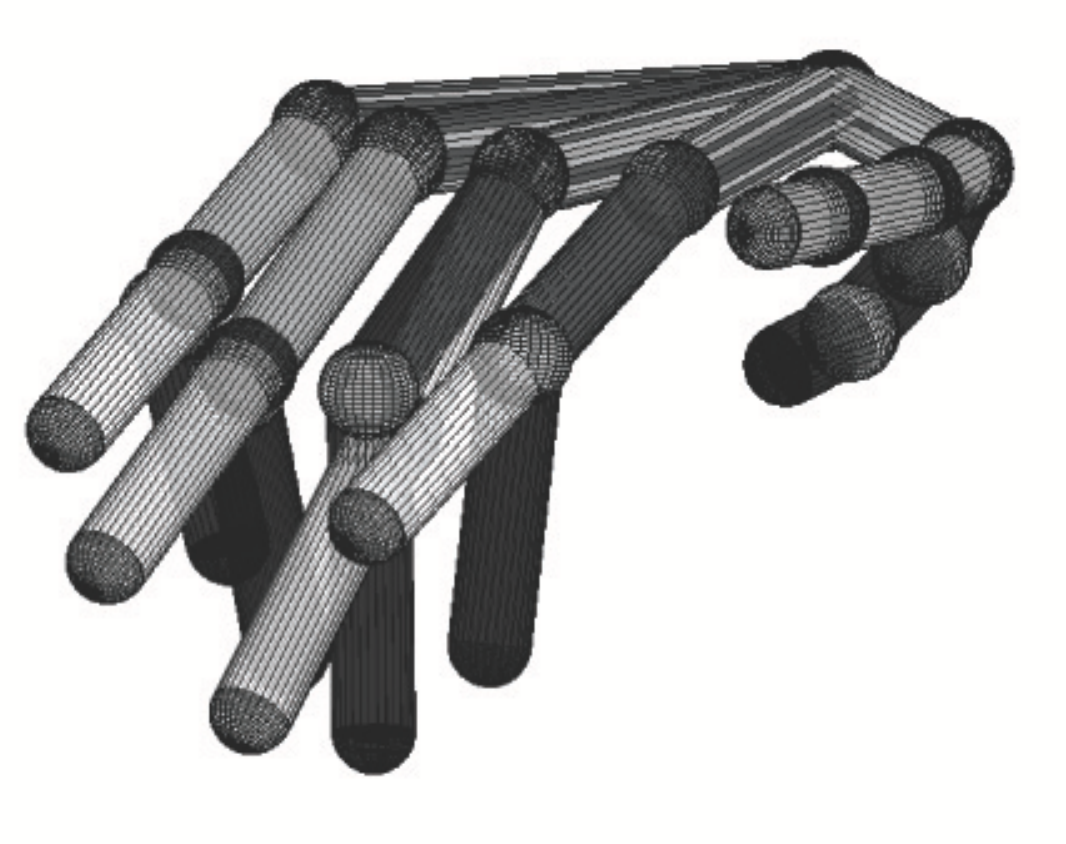} &
\includegraphics[width=0.2\textwidth]{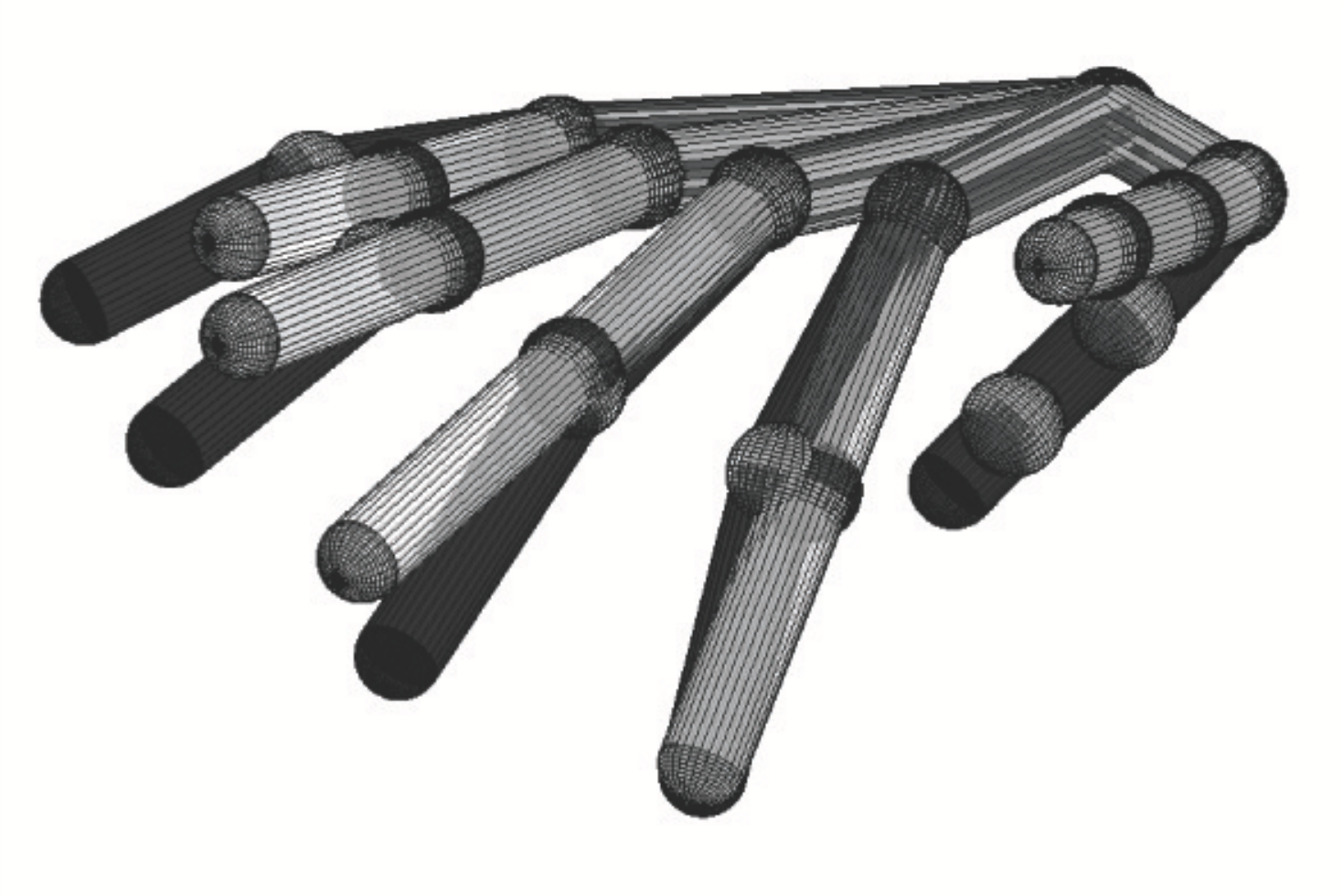}\\
\hline
{\rotatebox{90}{\mbox{MVE}}} &
\includegraphics[width=0.21\textwidth]{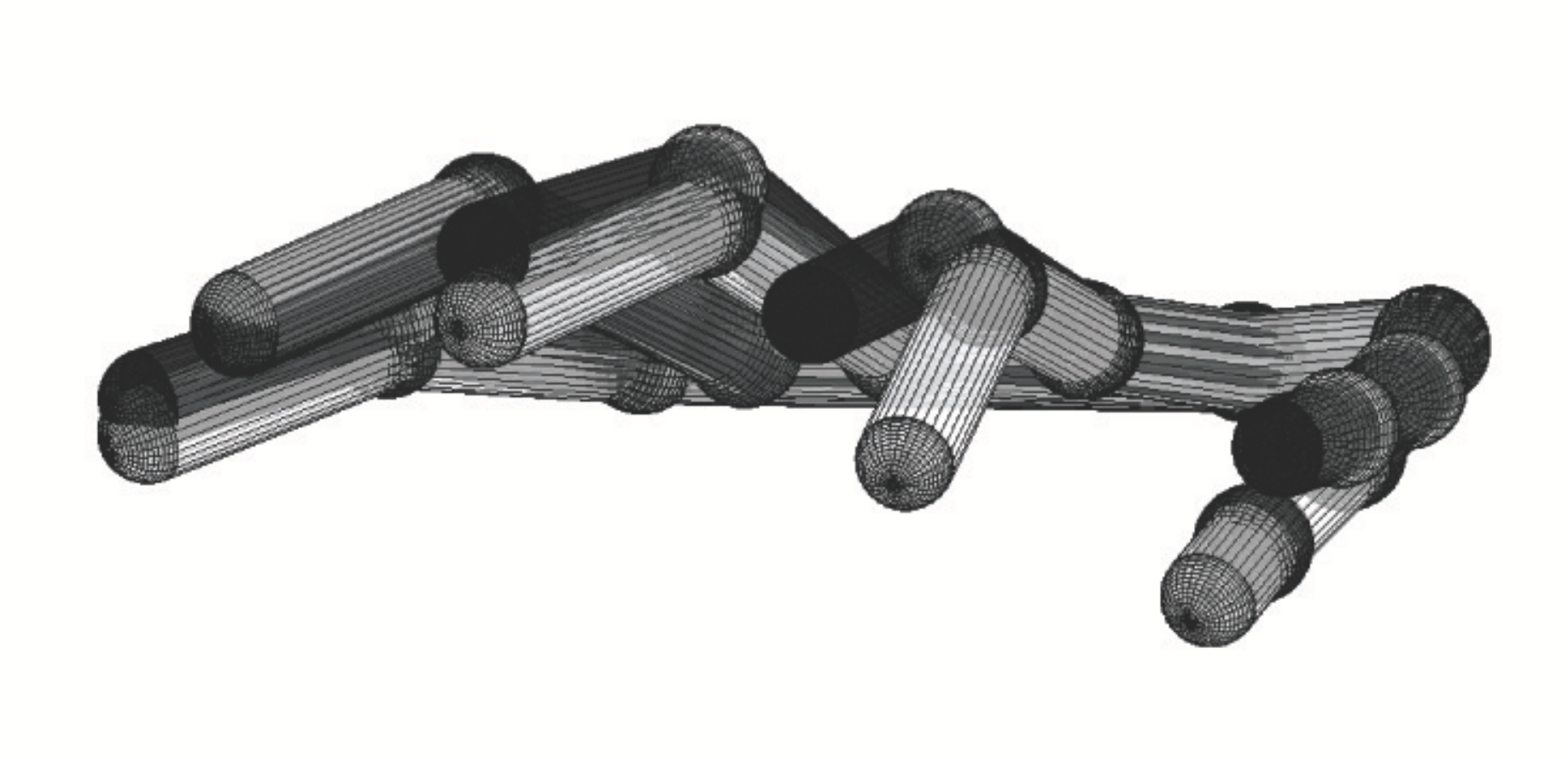} &
\includegraphics[width=0.21\textwidth]{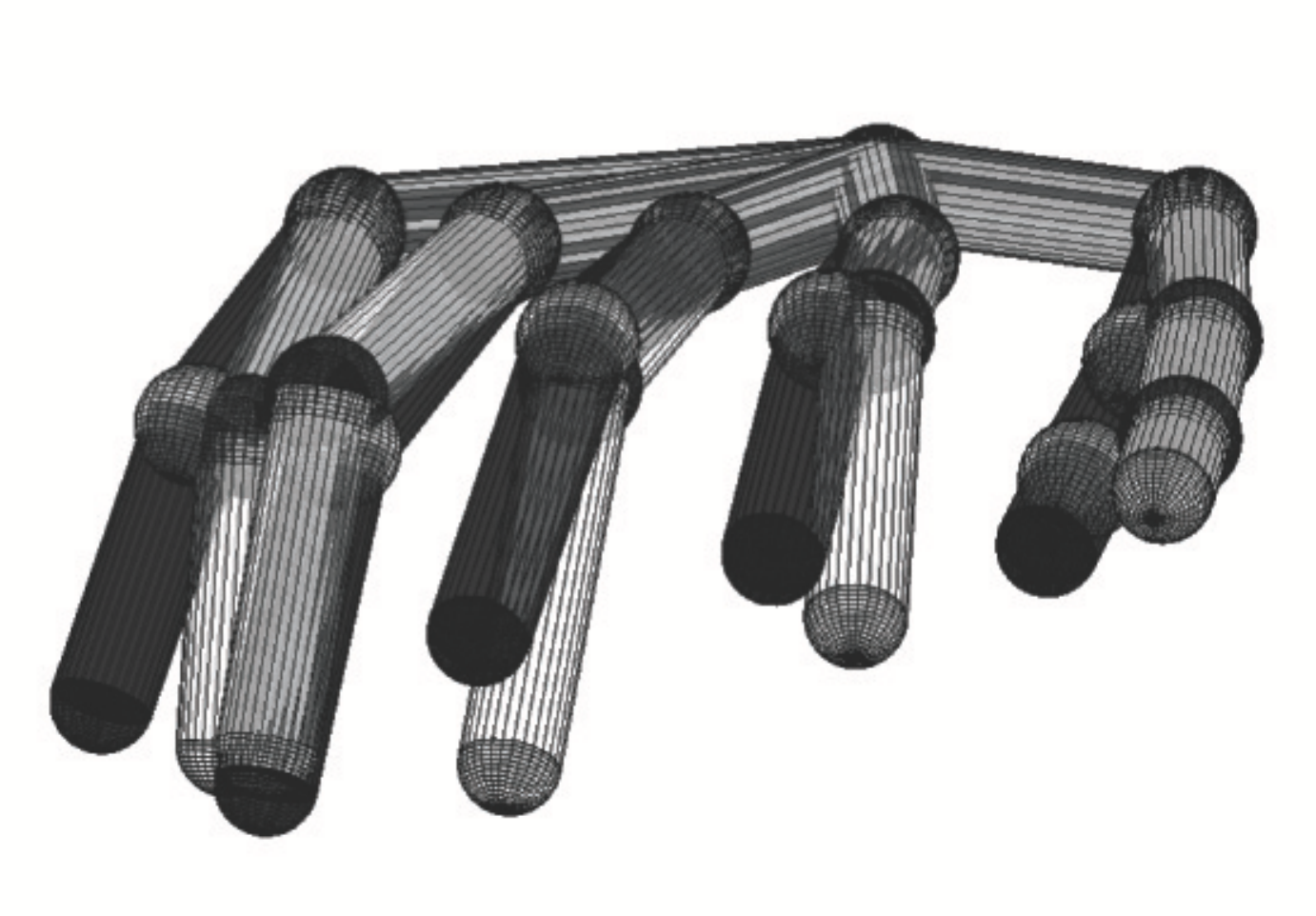} &
\includegraphics[width=0.18\textwidth]{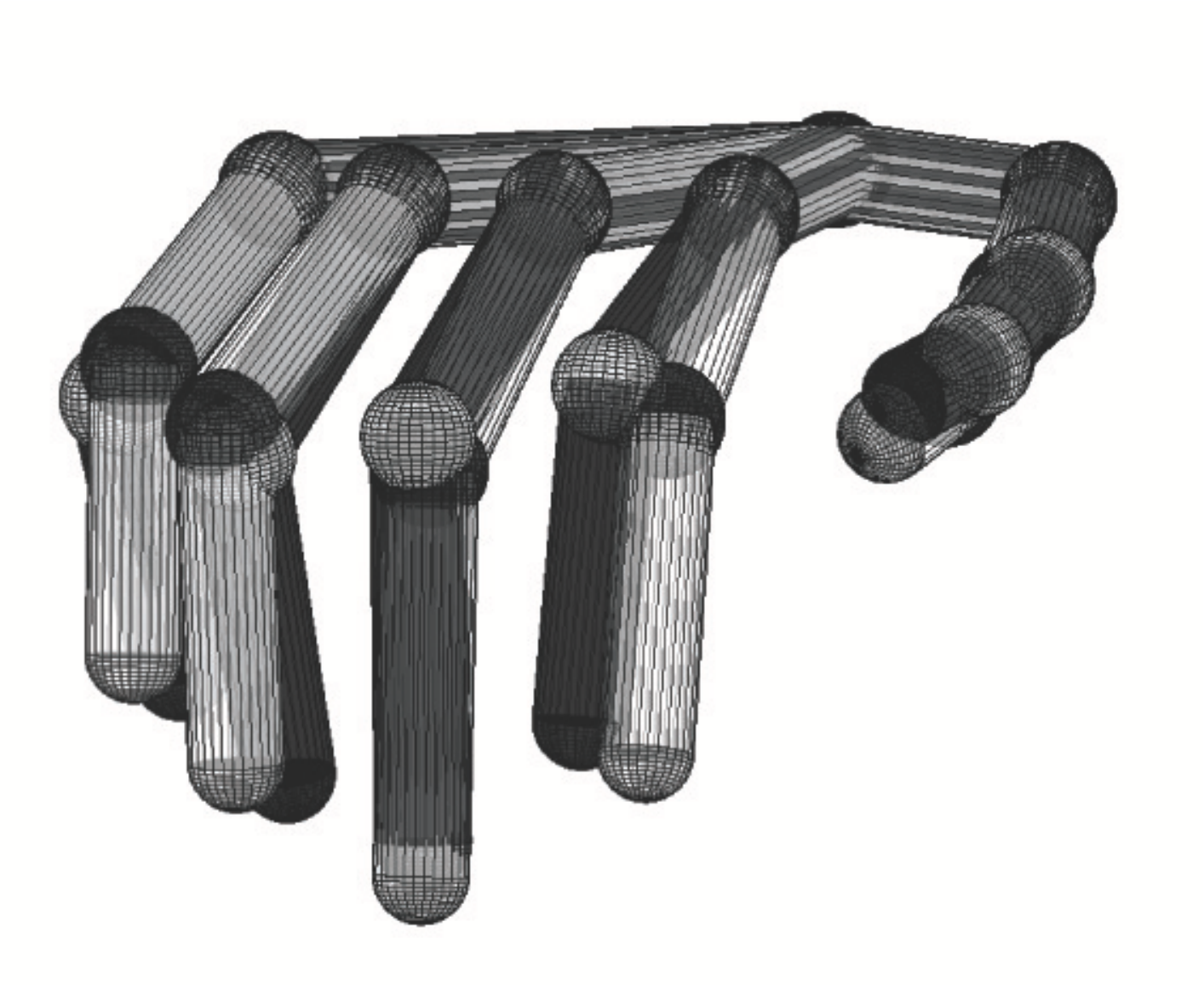} &
\includegraphics[width=0.2\textwidth]{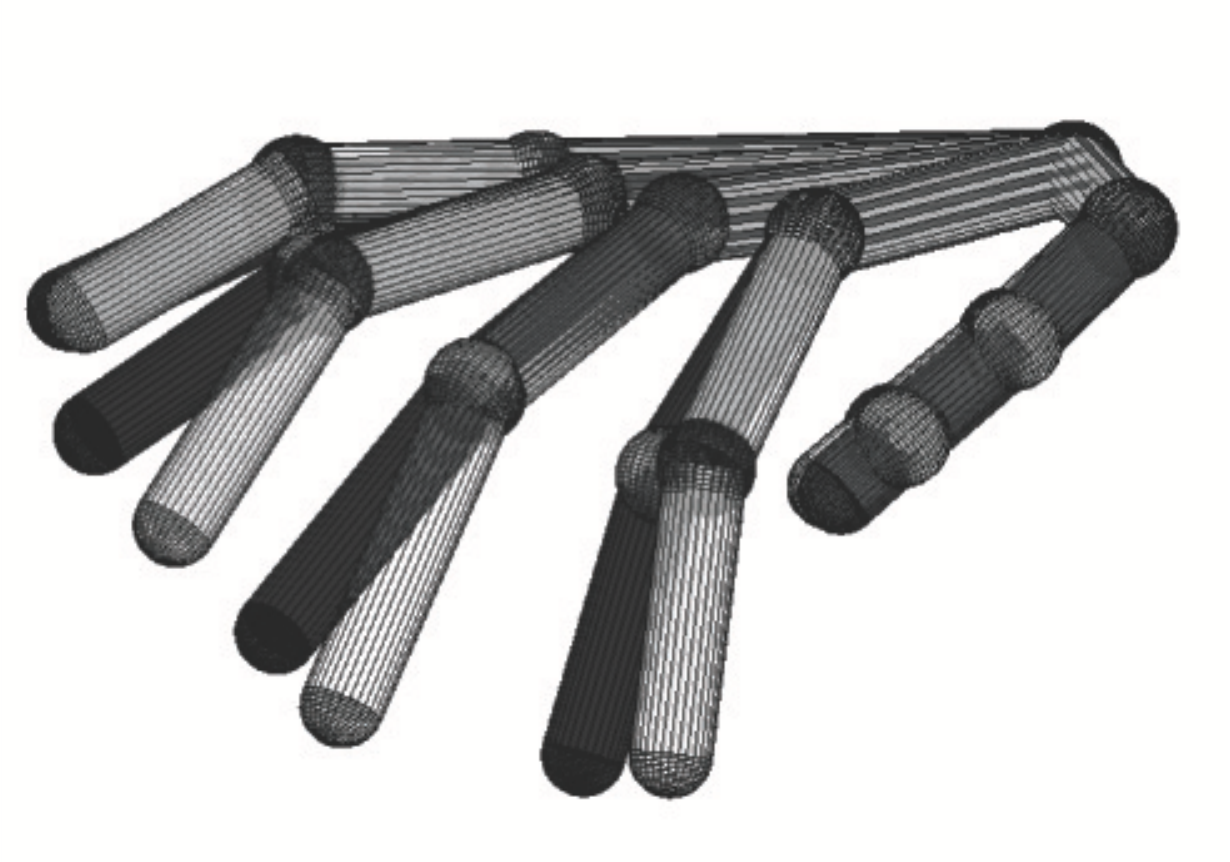}\\
\hline
\end{tabular}
\end{center}
\caption{Hand pose reconstructions with Pinv and MVE algorithms by using a selection matrix $H_s$ which allows to measure TM, IM, MM, RM and LM (see figure~\ref{fig:KinMod}). In color the ``real'' hand posture whereas in white the estimated one.}
\label{fig:PoseEst_WithAndWithoutNoise}
\end{figure*}

In figure~\ref{fig:PoseEst_WithAndWithoutNoise} some reconstructed poses are displayed in comparison with their corresponding reference values achieved with Phase Space System, with and without noise. Notice that MVE qualitatively shows the best reconstruction results, thus maintaining, unlike Pinv, the likelihood with common grasping poses because of the \emph{a priori} information.

\section{Experimental Results}
\label{Glove}
We test for the effectiveness of our reconstruction procedure using a sensorized glove based on Conductive Elastomer (CE). CE strip are printed on a Lycra$^\text{\textregistered}$\normalsize/cotton fabric in order to follow the contour of the hand, see figure~\ref{fig:guantol}. Connection to 20 different sensor segments of the polymeric strip is realized using additional conductive elastomer elements printed on the dorsal side of the glove~\cite{Tognetti}.

Since CE materials present piezoresistive characteristics, sensor elements corresponding to different segments of the contour of the hand length change as the hand moves. These movements cause variations in the electrical properties of the material, which can be revealed by reading the voltage drop across such segments.
The sensors are connected in series thus forming a single sensor line while the connections intersect the sensor line in the appropriate points. An \emph{ad hoc} electronic front-end was designed to compensate the resistance variation of the connections, made by the same material of the sensors, using an high input impedance stage.

Data coming from the front-end is then low pass filtered, digitalized and acquired by means of a general purpose DAQ card, and finally elaborated on a computer.

This glove can be considered as a continuous sensing device as it will be more clearly described in~\cite{Bianchi_etalII} since each sensor consists of a single sensing line all over different joints. However, data processing is based on the assumption that changes in the electrical characteristics of the sensor elements, corresponding to different segments of the contour of the hand, are associated with changes in the angle of the joint such sensor elements cut across. Furthermore, it was assumed that the hand aperture linearly relates to changes in the electrical characteristics of the sensor elements occurring as joint angles change. Considering this, a function that map the sensor raw data to joint angles was designed. Therefore, a linear regression model having hand aperture as dependent variable and the output of the sensor elements as independent variables was built. Fitting of the model was achieved using a calibration phase. Subjects were instructed to perform two fixed hand gesture, flat and grip, corresponding respectively to the minimum and maximum elongation of sensor's length.
In the present study, long finger flexion-extension recognition is obtained by means of an updated multi-regressive model having the metacarpophalangeal (MCP) flexion-extension angles of the five long fingers as dependent variables and the outputs of CE sensor covering MCP joints as independent ones. According to the hand kinematic model adopted in this work they are referred as to TM, IM, MM, RM, LM.  The model parameters are identified by measuring the sensor status in two different positions: (1) hand totally closed (90 degrees), (2) hand totally opened (0 degrees). For a more complete description of the aforementioned processing techniques as well as the glove design see e.g.~\cite{LorussiSJ2004,Tognetti2007,Tognetti2008}.

This sensorized glove represents one of the most recent and inexpensive envision in glove device literature.
However, this solution is limited by some factors, e.g.~cloth support which affects measurement repeatability as well as hysteresis and non linearities due to piezoresistive material properties. Moreover, the assumptions done for data processing (the relationship between joint angle and sensors as well the linearity between hand aperture and electrical property changes) as well the calibration phase, which is based only on two-point fitting, can act like potential sources of errors. To overcome the latter point a new calibration is performed to estimate the measurement matrix, as it is described in the next section.

\subsection{Results and Discussion}

Firstly we have obtained an estimation of the measurement matrix of the glove $H_g$. For this purpose, a calibration phase was performed by collecting a number of poses $N$ in parallel with the glove and the position optical tracking system. This number has to be larger or equal than the dimension of the state to estimate, i.e.~$N \geq 15$.  $X_g \in \real^{15\times15}$ collects the reference poses, while matrix $Y_g \in \real^{5\times15}$ organizes the measures from the glove. These measures represent the values of the signals referred to measured joints, averaged over the last 50 acquired samples. For the acquisition a DAQ card which works at 250 kS/s (NI PCI-6024E by National Instruments, Austin, Texas, USA) has been used within Matlab Simulink$^\text{\textregistered}$ environment.

An estimation $\hat{H}_g$ of the measurement matrix can be obtained by using the relation $Y_g=\hat{H}_gX_g$ as
\begin{equation}
\hat{H}_g=Y_g((X_g^{T})^{\dagger})^{T}\,.
\label{eq:jk}
\end{equation}

We characterize measurement noise in terms of fluctuations w.r.t.~the aforementioned average values of the measures, thus obtaining noise covariance matrix $R$. Noise level is less than 10\% measurement amplitude, however we might obtain consistent errors in the measurement matrix estimation due to intrinsic non-linearities and hysteresis of glove sensing elements.

The average absolute pose estimation error with MVE is $10.94\pm4.24^{\circ}$, while it is equal to $19.00\pm3.66^{\circ}$ by using Pinv. Statistical difference is observed between the two techniques ($\text{p}=0$, $T_{eq}$).
Notice that MVE exhibits the best pose reconstruction performances also in terms of maximum errors (25.18~$^{\circ}$ for MVE vs.~ 30.30~$^{\circ}$ for Pinv).
\begin{table}[t!]
\centering
\includegraphics[width=0.7\columnwidth]{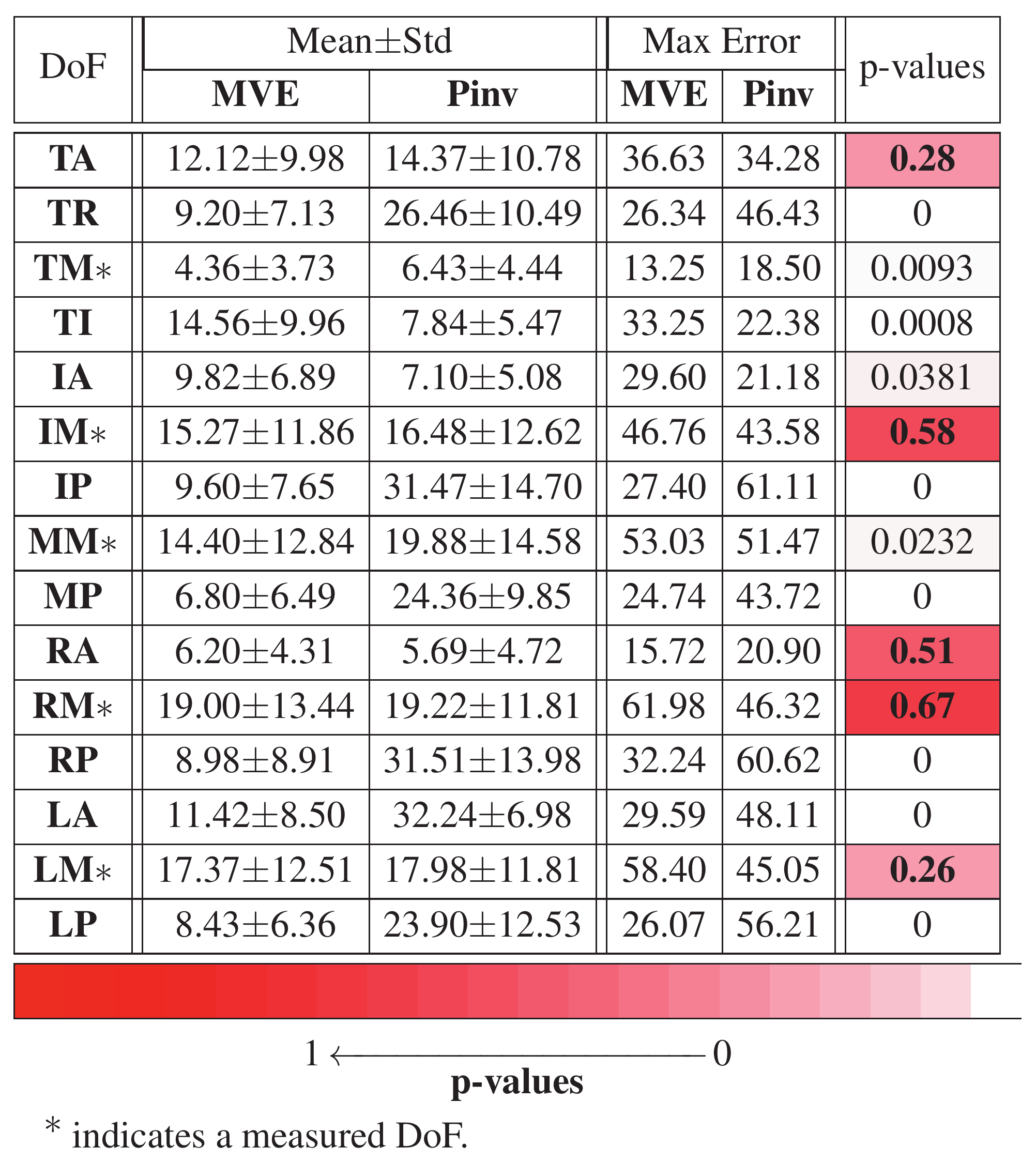}
\caption{Average estimation errors and standard deviations for each DoF $[^{\circ}]$, for the
sensing glove acquisitions. MVE and Pinv methods are considered. Maximum errors are also reported as well as p-values from the evaluation of DoF estimation errors between MVE and Pinv. A color map describing p-values is also added to simplify result visualization. $\diamond$ indicates  that $T_{eq}$ test has been exploited for the comparison. $\ddagger$ indicates a $T_{neq}$ test. When no symbol appears near the tabulated values, it means that $U$ test has been used. $\bold{Bold}$ value indicates no statistical difference between the two methods under analysis at 5\% significance level. When the difference is significative, values are reported with a $10^{-4}$ precision. p-values less than $10^{-4}$ are considered equal to zero.}
\label{tab:Giuntiguanto}
\end{table}
Absolute average reconstruction errors for each DoF are reported in table~\ref{tab:Giuntiguanto}. MVE produces smaller mean errors than those obtained with Pinv with statistical difference w.r.t.~Pinv algorithm, see table~\ref{tab:Giuntiguanto}, except, respectively, for those DoFs which are directly measured (i.e. IM, RM and LM), for RA DoF, which exhibits a limited average estimation error ($\approx 6^\circ$), and finally TA. For TI the smallest average estimation is observed with Pinv; a possible explanation for this might be still related to the difficulties behaviour related to thumb phalanx modeling (as also observed in section~\ref{Section:mod} for TA DoF) can be explained considering the simplifications adopted in our kinematic model for modeling the complexity of thumb phalanx (as observed also in Section~\ref{Section:mod} for TA DoF). IA DoF presents the smallest absolute average estimation error with Pinv, although p-values from the comparisons between the two techniques for the estimation of this DoF are close to the significance threshold. All these observations are coherent with the discussions developed in Section~\ref{Section:mod}.

Maximum DoF reconstruction errors for MVE are observed especially for those measured DoFs with potentially maximum variations in grasping tasks; this fact may be probably interpreted considering the non linearities in sensing glove elements leading to inaccurate estimation of $H_g$, hence to inaccurate measures.

\begin{figure*}[t!]
\begin{center}
	\renewcommand{\arraystretch}{1.5}
\begin{tabular}[c]{c}
Real Hand Postures\\
\end{tabular}
\begin{tabular}[c]{p{0.3cm}|p{2.4cm}|p{2.4cm}|p{2.4cm}|p{2.4cm}|}
\cline{2-5}
 &
\includegraphics[width=0.2\textwidth]{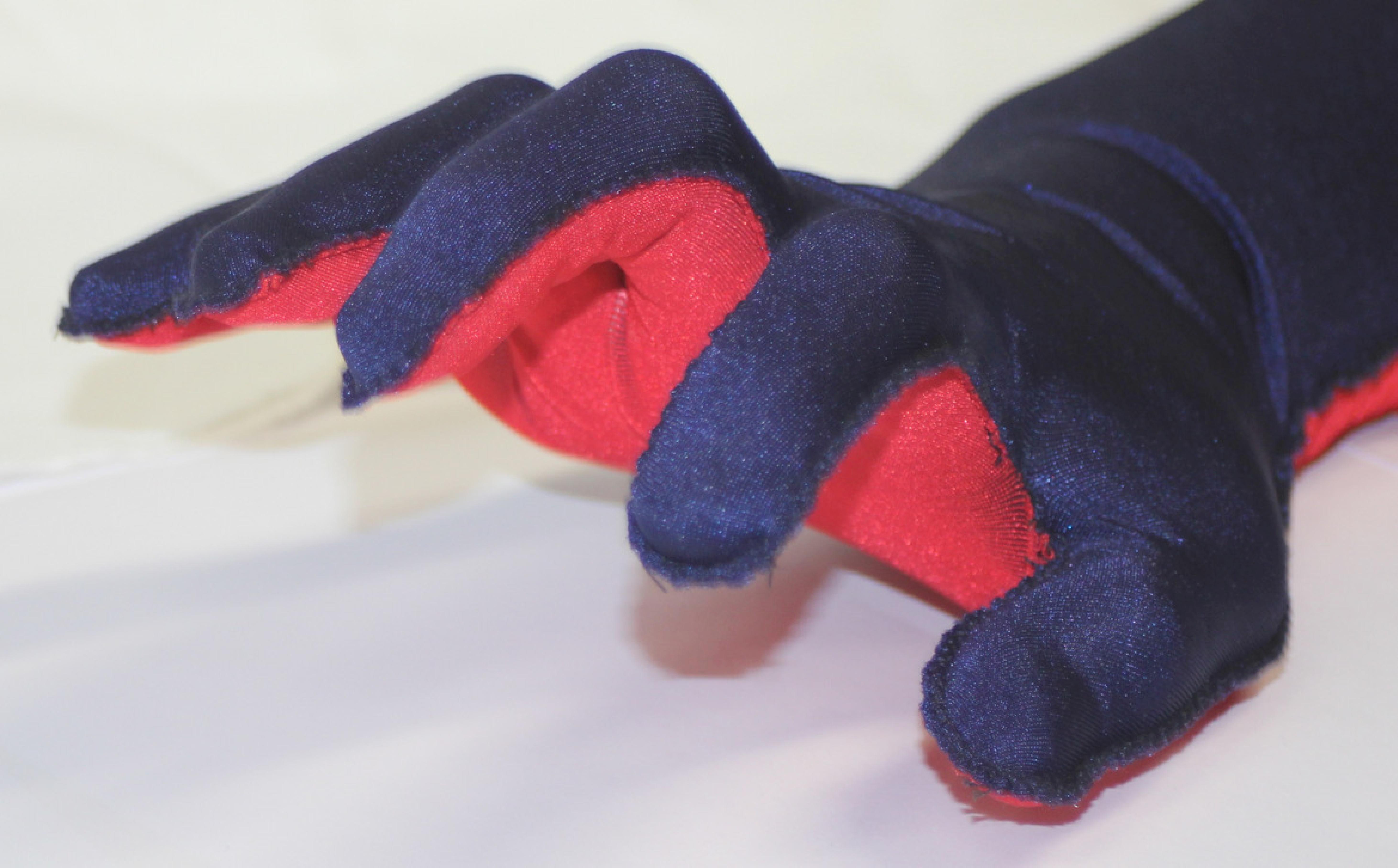} &
\includegraphics[width=0.18\textwidth]{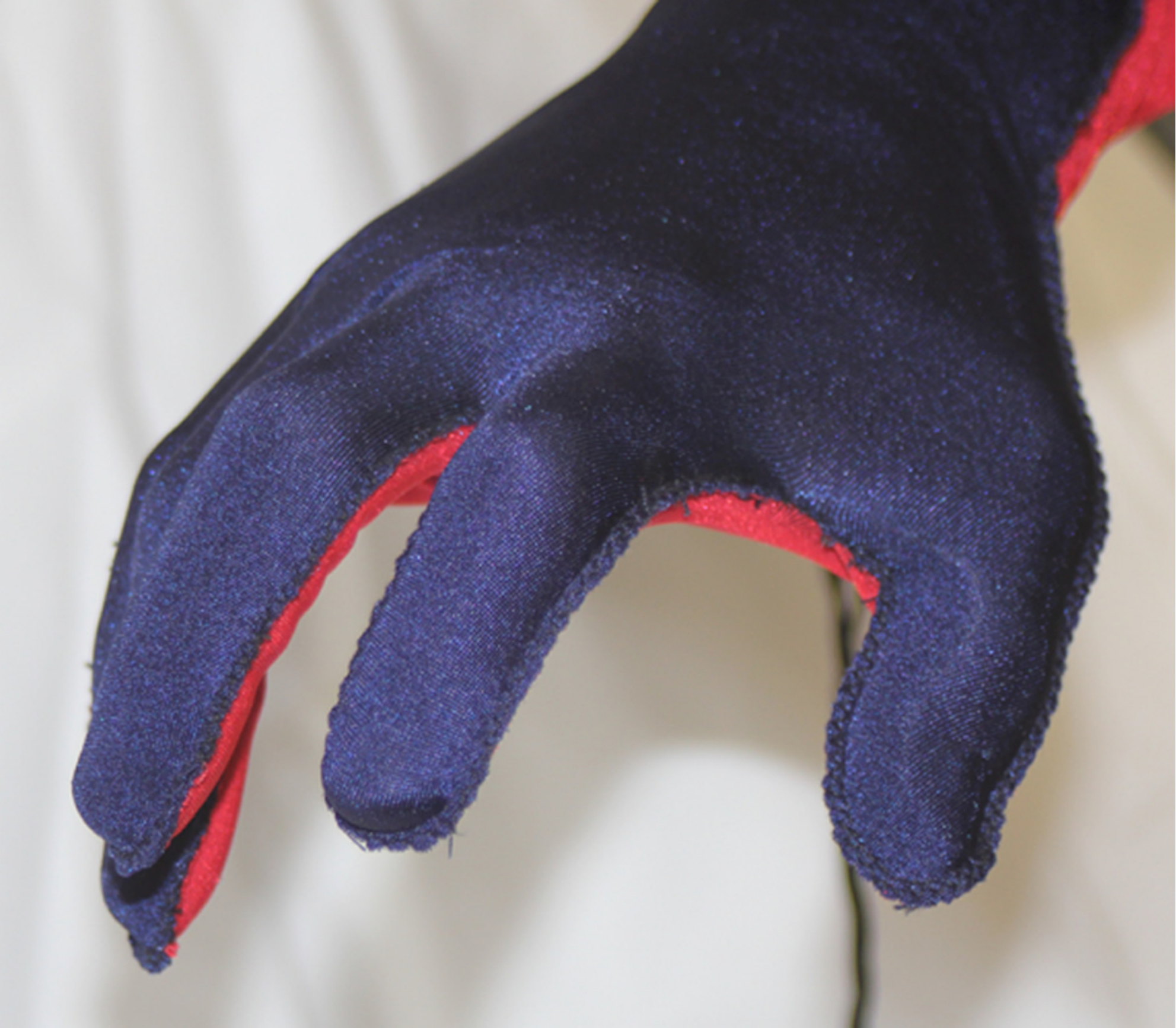} &
\includegraphics[width=0.18\textwidth]{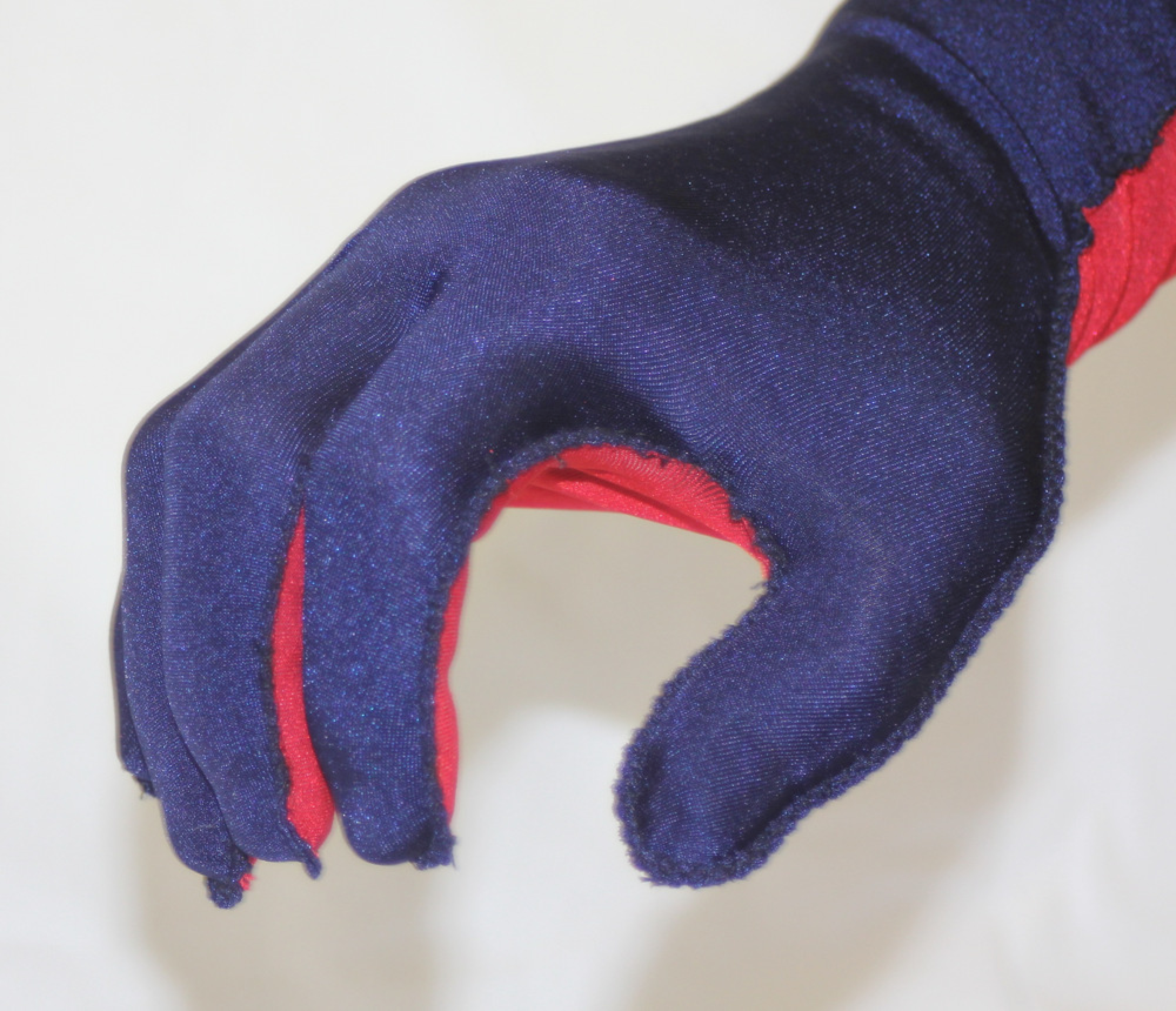} &
\includegraphics[width=0.2\textwidth]{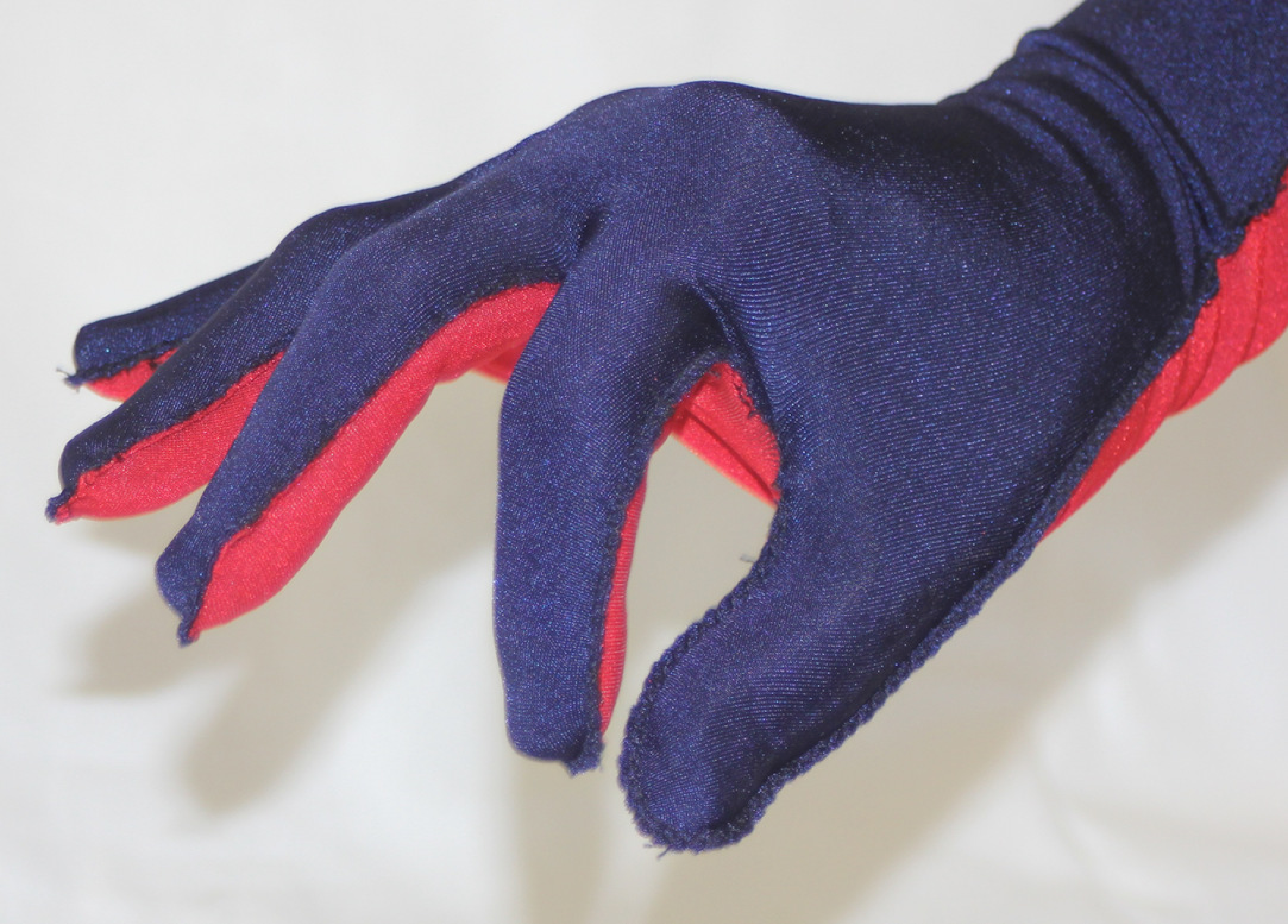}\\
\cline{2-5}
\hline
\end{tabular}
\begin{tabular}[c]{|p{0.3cm}|p{2.4cm}|p{2.4cm}|p{2.4cm}|p{2.4cm}|}
{\rotatebox{90}{\mbox{Pinv}}} &
\includegraphics[width=0.21\textwidth]{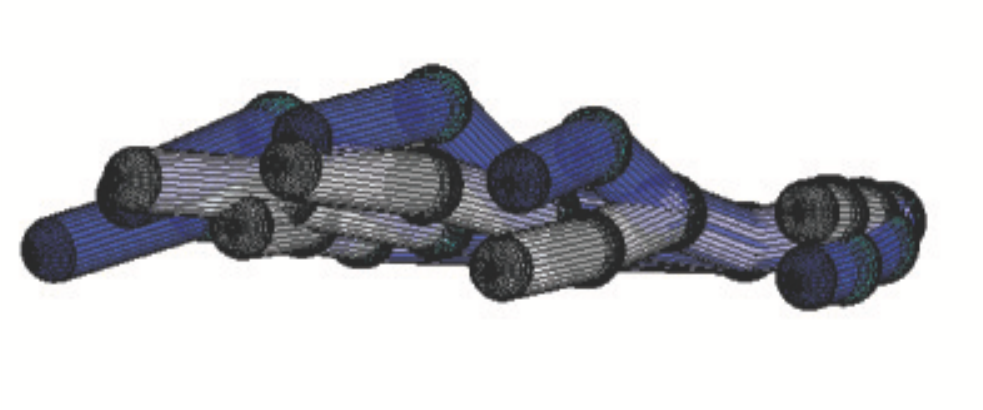} &
\includegraphics[width=0.2\textwidth]{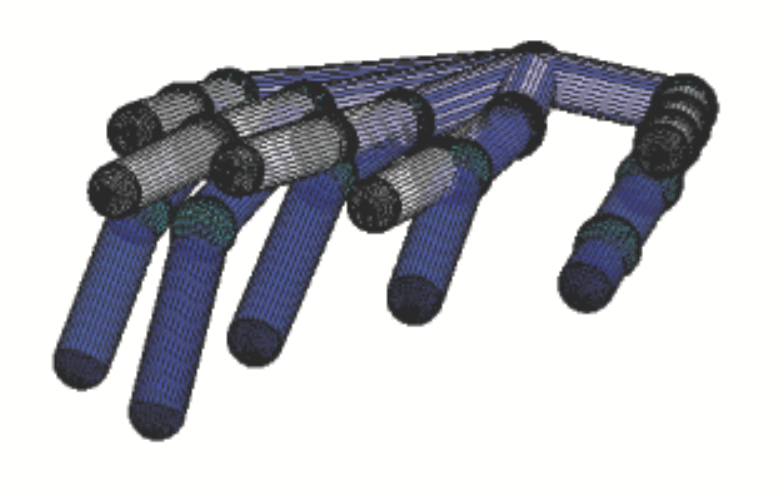} &
\includegraphics[width=0.2\textwidth]{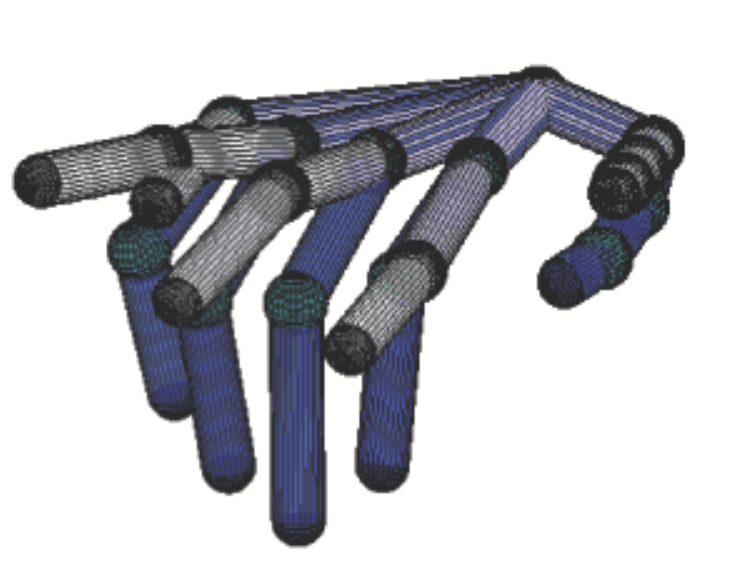} &
\includegraphics[width=0.2\textwidth]{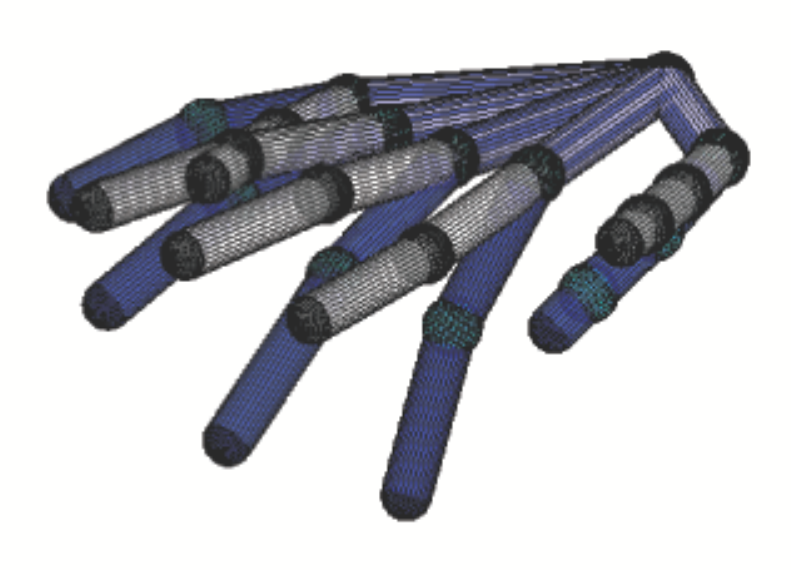}\\
\hline
{\rotatebox{90}{\mbox{MVE}}} &
\includegraphics[width=0.21\textwidth]{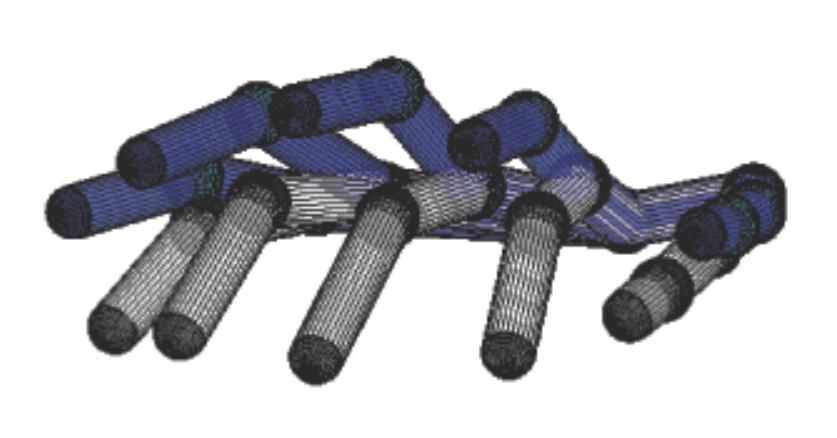} &
\includegraphics[width=0.2\textwidth]{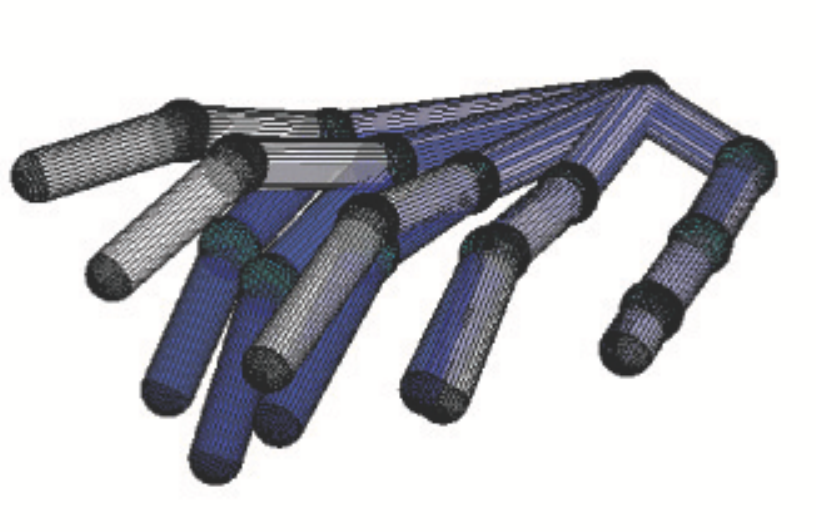} &
\includegraphics[width=0.18\textwidth]{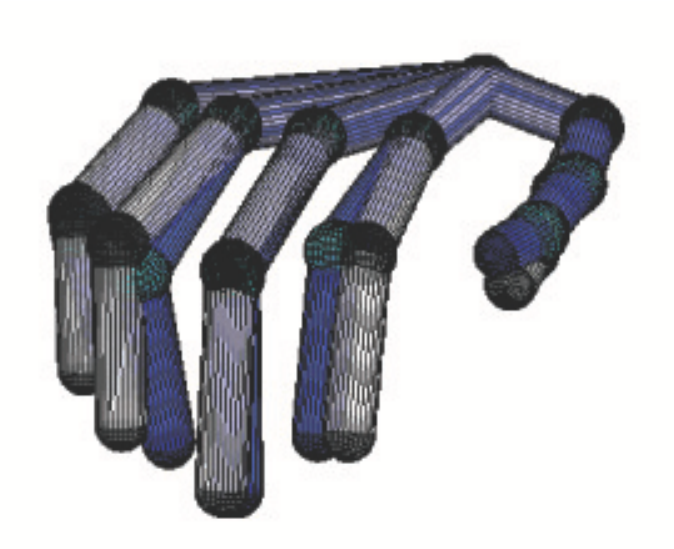} &
\includegraphics[width=0.2\textwidth]{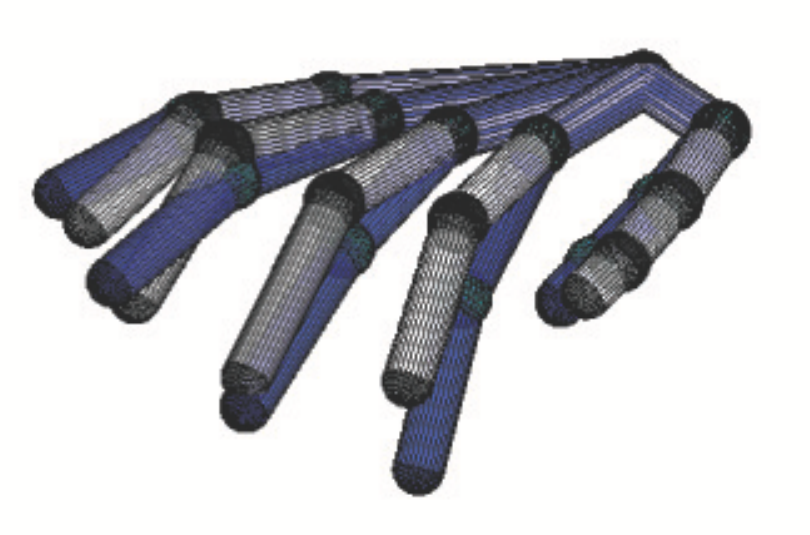}\\
\hline
\end{tabular}
\end{center}
\caption{Hand pose reconstructions with Pinv and MVE algorithms, with measures given by sensing glove. In blue the ``real'' hand posture whereas in white the estimated one.}
\label{fig:RecG}
\end{figure*}

Finally, except for some singular poses, the best estimation performance is provided by MVE for which a good robustness to errors in measurement process modeling is also observed. However, the latter errors have not been taken numerically into account in our analyses. Moreover, as it can been seen in figure~\ref{fig:RecG}, reconstructed hand configurations obtained by MVE preserve likelihood with real poses, as opposed to pseudo-inverse based algorithm.

\section{Conclusions}

In this work reconstruction techniques to estimate static hand poses from a reduced number of measures given by an input glove-based devices are presented. These techniques are based on classic optimization and applied optimal estimation methods. The main innovation relies on the exploitation of the \emph{a priori} information embedded in the covariance structure of a set of grasp poses. This covariance individuates some coordination patterns, defined as \emph{postural synergies}, which reduce hand DoFs to be measured and controlled.

Simulations results, where noise effects are also considered, and experiments with a low-cost sensing glove are reported. Performance is compared with the one obtained with a simple pseudo-inverse based algorithm. Statistical analyses demonstrate the effectiveness of the here proposed hand pose reconstructions.

The achieved results can be useful to improve a large class of human-interfaces in many application fields, e.g.~video-games or tele-robotics, where fine hand position individuation and low cost devices are crucial features to allow a reliable haptic experience.

In~\cite{Bianchi_etalII} we apply this reconstruction procedure to the measures provided by an optimally designed sensing glove.

\section*{Acknowledgment}
Authors gratefully acknowledge Alessandro Tognetti and Nicola Carbonaro for their assistance in sensing glove configuration setting and Armando Turco for his help in data acquisition with the optical tracking system.

\bibliographystyle{apalike}

\end{document}